\documentclass[letterpaper, 10 pt, conference]{ieeeconf}  

\IEEEoverridecommandlockouts                              

\usepackage{amsmath} 
\usepackage{amssymb}  

\usepackage{graphicx}
\usepackage[tight,footnotesize]{subfigure}
\usepackage{amsmath} 
\usepackage{amssymb}  
\usepackage{hyphenat}
\usepackage[table]{xcolor}
\usepackage{caption}
\usepackage{multirow}
\usepackage{url}
\usepackage{booktabs,amsfonts,dcolumn}
\usepackage{tabularx}
\usepackage{gensymb}
\usepackage{makecell}
\usepackage{cite}
\usepackage[group-separator={,}]{siunitx}
\usepackage{tikz}

\urldef{\mailmd}\path| md.modasshir@outlook.com|
\urldef{\mailsb}\path| yiannisr@cse.sc.edu|

\usepackage{xspace}

\long\def\invis#1{}

\newcommand\eq[1]{Eq.~\eqref{#1}}
\newcommand\fig[1]{Fig.~\ref{#1}}

\newcommand\tab[1]{Table~\ref{#1}}

\makeatletter
\DeclareRobustCommand\onedot{\futurelet\@let@token\@onedot}
\def\@onedot{\ifx\@let@token.\else.\null\fi\xspace}

\def\etal{\emph{et al}\onedot}
\makeatother

\newcommand\copyrighttext{%
  \footnotesize \textcopyright 2020 IEEE. Personal use of this material is permitted.
  Permission from IEEE must be obtained for all other uses, in any current or future
  media, including reprinting/republishing this material for advertising or promotional
  purposes, creating new collective works, for resale or redistribution to servers or
  lists, or reuse of any copyrighted component of this work in other works.}
\newcommand\copyrightnotice{%
\begin{tikzpicture}[remember picture,overlay]
\node[anchor=south,yshift=10pt] at (current page.south) {\fbox{\parbox{\dimexpr\textwidth-\fboxsep-\fboxrule\relax}{\copyrighttext}}};
\end{tikzpicture}%
}

\title{\LARGE \bf
DeepURL:\\
Deep Pose Estimation Framework for Underwater Relative Localization 
}
\author{Bharat Joshi$^a$$^*$, Md Modasshir$^a$$^*$, Travis Manderson$^b$, Hunter Damron$^a$, Marios Xanthidis$^a$,\\ Alberto Quattrini Li$^c$, Ioannis Rekleitis$^a$, Gregory Dudek$^b$%
\thanks{$^a$University of South Carolina, Columbia, SC, USA, 29208, {\tt\small \{bjoshi,modasshm,mariosx,hdamron\}@email.sc.edu, yiannisr@cse.sc.edu}. The authors would like to acknowledge the generous support of the National Science Foundation grants (NSF 2024741, 1943205)}. %
\thanks{$^b$McGill University, Montreal, Quebec, Canada,  {\tt\small \{travism,dudek\}@cim.mcgill.ca}}%
\thanks{$^c$Dartmouth College, Hanover, NH, USA, 03755 {\tt\small alberto.quattrini.li@dartmouth.edu}}%
\thanks{$^*$First two authors contributed equally.}
}
\begin{document}
\maketitle              
\copyrightnotice
\thispagestyle{empty}
\pagestyle{empty}

\begin{abstract}
    
In this paper, we propose a real-time deep learning approach for determining the 6D relative pose of Autonomous Underwater Vehicles (AUV) from a single image. A team of autonomous robots localizing themselves in a communication-constrained underwater environment is essential for many applications such as underwater exploration, mapping, multi-robot convoying, and other multi-robot tasks. Due to the profound difficulty of collecting ground truth images with accurate 6D poses underwater, this work utilizes rendered images from the Unreal Game Engine simulation for training. An image-to-image translation network is employed to bridge the gap between the rendered and the real images producing synthetic images for training. The proposed method predicts the 6D pose of an AUV from a single image as 2D image keypoints representing 8 corners of the 3D model of the AUV, and then the 6D pose in the camera coordinates is determined using RANSAC-based PnP. Experimental results in real-world underwater environments (swimming pool and ocean) with different cameras demonstrate the robustness and accuracy of the proposed technique in terms of translation error and orientation error over the state-of-the-art methods. The code  is publicly available.\footnote{\url{https://github.com/joshi-bharat/deep_underwater_localization}}

\invis{A team of autonomous robots localizing themselves, in a communication-constrained underwater environment, is essential for many applications such as underwater exploration, mapping, multi-robot convoying, and other multi-robot tasks. In this paper, we propose a deep-learning based method for determining the 6D relative pose of Autonomous Underwater Vehicles from a single image. \invis{The underwater domain presents a significant challenge in collecting images of the robot with corresponding accurate 6D poses. To overcome this limitation, the proposed method is trained on data generated from an Unreal Game Engine simulation but tested on real datasets gathered from field deployments.} We utilize an image translation network to bridge the gap between the rendered with the real datasets, while training on CycleGAN generated synthetic images with soft labels. The proposed method, at first, employs an object detection based, 6D pose estimation framework, where patches inside object detection box contribute to local pose prediction in the form of 2D keypoint locations. Then, the predicted confidences are utilized to combine these pose coordinates into 3D\hyp to \hyp2D correspondences. Finally, the relative pose from the camera is found using RANSAC-based PnP. Experimental results from different deployments of two AUVs demonstrates real-time performance and the robustness of the proposed technique in a variety of real underwater scenarios.}
\end{abstract}

\section{INTRODUCTION}
The ability to localize is crucial to many robotic applications. There are several environments where keeping track of the vehicle's position is a challenging task; particularly in GPS\hyp denied environments with limited features. A common approach to address the localization problem is to use intra\hyp robot measurements for improved positional accuracy -- an approach termed Cooperative Localization (CL)~\cite{Rekleitis1998}.  Central to CL is the ability to estimate the relative pose between the two robots; this estimate can then be utilized to improve the absolute localization based on the global pose estimates for one of the two robots. A video overview is also available online$^1$.

In this paper, we propose and evaluate a deep  pose  estimation framework for underwater relative localization, called \emph{DeepURL}. The primary application motivating this work is underwater exploration and mapping by a team of autonomous underwater vehicles (AUVs)   with a focus on shipwreck and underwater cave mapping; environments that are challenging to most existing localization methodologies (e.g., visual and visual/inertial\hyp based systems~\cite{quattrinili2016iser-vo,JoshiIROS2019}). Other applications include convoying~\cite{ShkurtiIROS2017}, environmental assessments~\cite{Manderson2017jfr}, informative navigation~\cite{Manderson2020rss}, and inspections. 

\begin{figure}[t]
    \begin{center}
    {\includegraphics[width=0.85\linewidth]{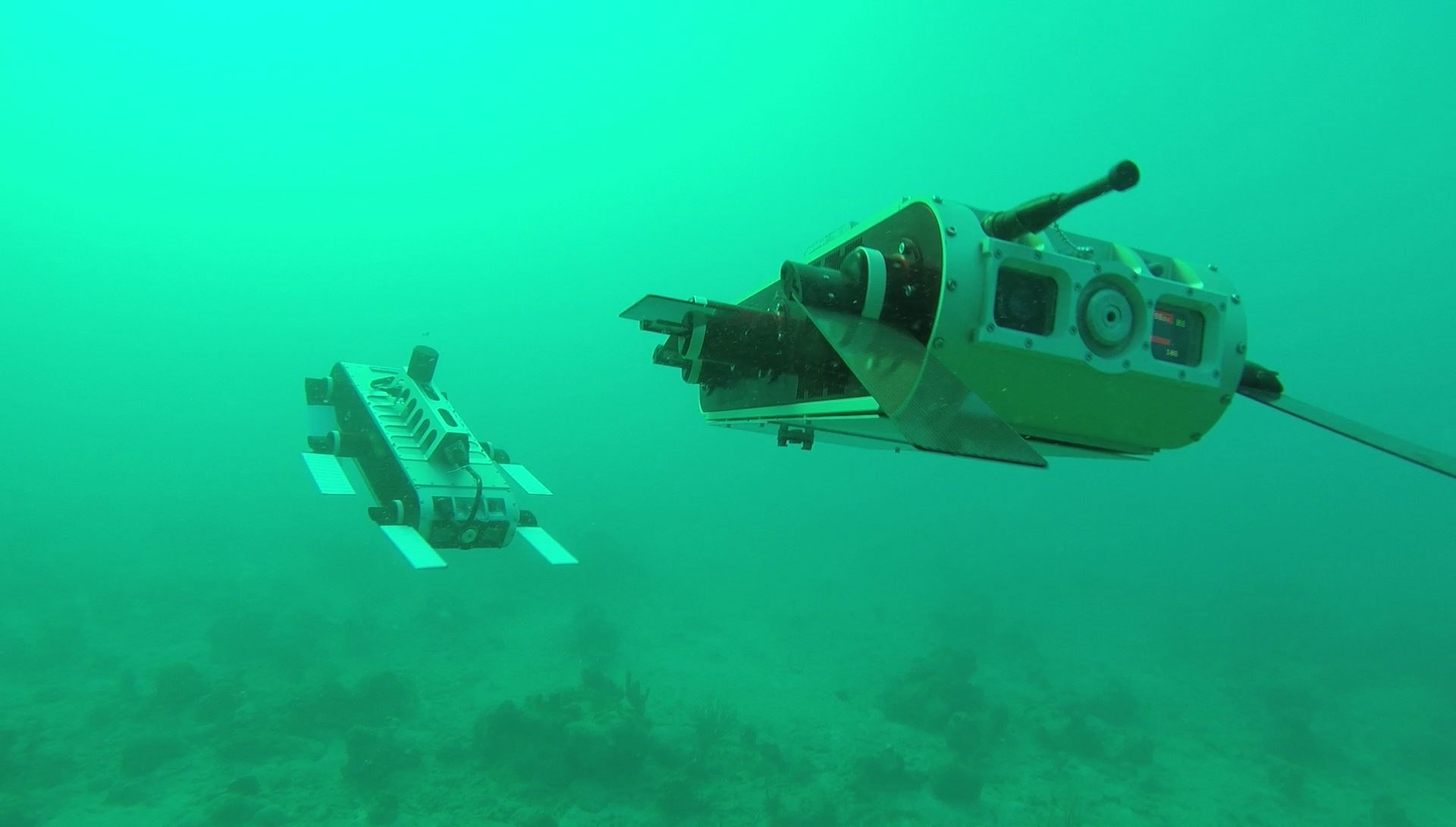}}
    \caption{Two Aqua2 vehicles collecting images over a reef require relative localization to efficiently cover the area.}
    \label{fig:beauty}
    \end{center}
\end{figure}

The proposed methodology draws from the rich object detection research and is adapted to the unique conditions of the underwater domain. Traditionally, 6D pose estimation (3D position and 3D orientation) is performed by matching feature points between 3D models and images~\cite{lowe1999object,rothganger20063d,collet2011moped, pose_tracking_ismar}. While these methods are robust when objects are well textured, they perform poorly when objects are featureless or textureless. In the underwater domain, particulates in the water generate undesired texture smoothing. Recent approaches~\cite{ssd_6d,Sundermeyer_2018_ECCV, xiang2017posecnn,bb8, single6dpose, segpose} to estimate 6D poses using deep neural network perform well on standard benchmark pose estimation datasets such as LINEMOD~\cite{linemod}, Occluded\hyp{LINEMOD}~\cite{occluded_linemod}, and YCB\hyp{Video}~\cite{xiang2017posecnn}, but they require either intensive manual annotation or a motion capture system. To the authors' knowledge, a readily applicable method for collecting underwater training data with the corresponding accurate 6D poses is not available. In this work, we focus on estimating the 6D pose of an Aqua2 vehicle~\cite{DudekIROS2005} (shown in Fig. \ref{fig:beauty}). The observer is either another Aqua2 robot or an underwater handheld camera. The proposed method utilizes the Unreal Engine 4 (UE4)~\cite{manderson2018aqua} with a 3D model of the Aqua2 robot swimming, projected over underwater images to generate training images with known poses for the pose estimation network. Dissimilarity in images arising from intrinsic factors such as distortion differences from different cameras, external factors such as color-loss, poor visibility quality, or the surroundings, hampers the performance of classical deep learning\hyp based 6D pose estimation methods in the underwater domain. CycleGAN~\cite{CycleGAN2017} was employed to transform UE4 rendered images to image sets used for training with varying in appearance, similar to real-world underwater images. 

Using a modified version of YOLOv3~\cite{yolov3} to detect an object bounding box, the proposed network produces robust 6D pose estimates by combining multiple local predictions of 2D keypoints that are projections of 3D corners of the object. Only grid cells inside the detected bounding box contribute to the selection of 2D keypoints along with a confidence score. Using the predictions with confidence, the most dependable 2D keypoint candidates for each 3D keypoint are selected to yield a set of 2D-to-3D correspondences. These selected 2D keypoints are used in the RANSAC-based PnP~\cite{epnp} algorithm to obtain a robust 6D pose estimate. 

The proposed framework has been tested in different environments -- pool, ocean -- and different platforms, including an Aqua2 robot and GoPro cameras, demonstrating its robustness. The main contributions are as follows:

\begin{itemize}
    \item A 6D pose prediction network that predicts object bounding boxes and eight keypoints in image coordinates. These 2D keypoints are then used in 2D\hyp to\hyp 3D correspondence to estimate 6D pose.
    
    \item We demonstrate the effective use of rendered image augmentation\footnote{Traditionally Generative Adversarial Networks (GAN)~\cite{CycleGAN2017} use the term image translation for this operation, however, the term translation can be confusing for a robotics application.} in 6D pose prediction, eliminating the need for ground truth labeling in real images. Utilizing image augmentation from the rendered to the underwater environment, the pose prediction network becomes invariant to color-loss, texture-smoothing, and other domain-specific challenges.

    \item We publish a dataset of the Aqua2 robot captured in the ocean and swimming pool to further research in the underwater domain\footnote{\url{https://afrl.cse.sc.edu/afrl/resources/datasets/}}.
\end{itemize}

The next section reviews related works. Section~\ref{sec:proposed} introduces the proposed method, including the synthetic data generated for training, and the pose estimation method.  Section~\ref{sec:experiments} presents first the ground truth data acquisition used exclusively for testing, then quantitative results from different datasets together with a comparison with other methods are discussed. Finally, we conclude the paper with future work in section \ref{sec:conclusions}.  
\begin{figure*}[ht]
\begin{center}
\vspace{0.1in}
\includegraphics[width=0.9\textwidth, ]{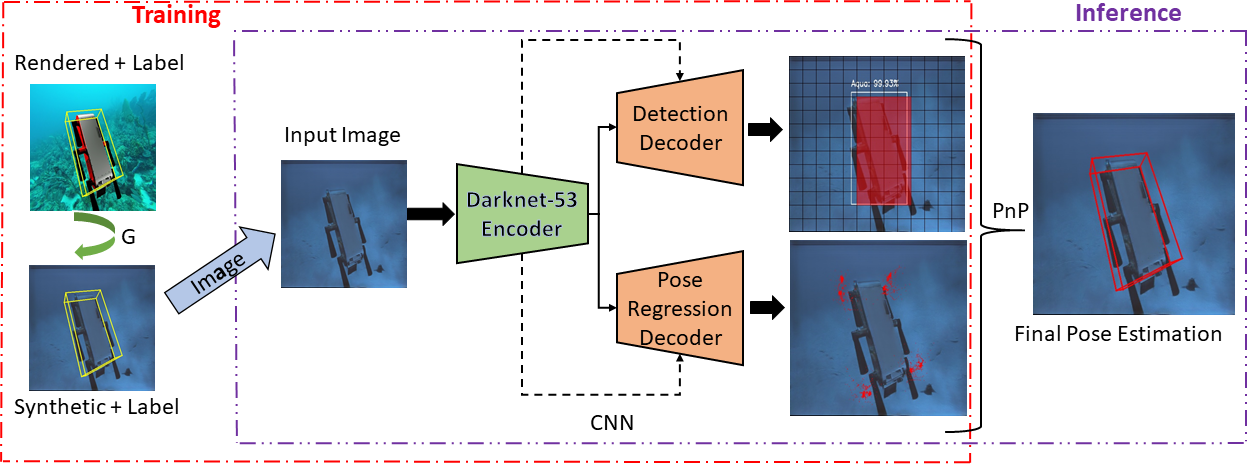}
\caption{In training (outlined in red), the rendered images are translated to the synthetic images resembling Aqua2 swimming in a pool or a ocean environment. The synthetic images are then fed to a common encoder, which is connected to two decoder streams: Detection Decoder (object detection) and Pose Regression Decoder (6D pose regression). Only in inference (outlined in purple), are the predicted 2D keypoint projections of 8 corners of the 3D Aqua2 model processed and utilized to obtain a 6D pose using the RANSAC-based PnP algorithm.}
 \label{fig:deepcl_pipeline}
 \end{center}
\end{figure*}

\section{RELATED WORK}\label{sec:related}
In this paper, we focus on 6D pose estimation using RGB images without access to a depth map.  RGB-D based methods~\cite{Brachmann2014,CHOI2016595,Sock2017Multiview6O} are not applicable underwater given the attenuation of infrared light at a very short distance. The classical approach for 6D object pose estimation involves extracting local features from the input image, matching them with features from a 3D model to establish 2D-to-3D correspondences from which a 6D pose can be obtained through the PnP algorithm~\cite{rothganger20063d,kostas_6D,lowe1999object,pose_tracking_ismar}. Previous work studied local feature descriptors invariant to changes in scale, rotation, illumination, and viewpoints~\cite{sift,NIPS2012_4848}. Even though these feature\hyp based techniques can handle occlusions and scene clutter, they require sufficient texture to compute local features. To deal with poorly-textured objects, some efforts focused on learned feature descriptors using machine learning techniques~\cite{Wohlhart_2015_CVPR, doumanoglou2016siamese}.

In recent years, pose estimation research has been dominated by frameworks utilizing deep neural network. These methods can be broadly  classified into two categories: either regressing directly to 6D pose estimates~\cite{ssd_6d,xiang2017posecnn}, or predicting 2D projections of 3D keypoints on an image and then obtaining pose via PnP algorithm~\cite{bb8, single6dpose}. Xiang \etal~\cite{xiang2017posecnn} estimate the object center in the image with the distance of the center used for estimating the translation and the predicted quaternions for object rotations. Peng \etal~\cite{peng2019pvnet} used a pixel-wise voting network to regress pixel-wise unit vectors pointing to the keypoints and used these vectors to vote for keypoint locations using RANSAC. Recent works ~\cite{Zakharov_2019_ICCV,Gupta_2019_ICCV_Workshops,li2017deepim}  researched on post-processing to refine the initial pose estimates from the first step. Li \etal~\cite{Li_2019_ICCV} disentangled
the pose to predict rotation and translation separately from two different branches to increase accuracy. 

Recent approaches \cite{ipose,Oberweger2018MakingDH} for pose estimation focus on local patches belonging to the object rather than producing a single global prediction. The work of Hu \etal~\cite{segpose} is closest to our approach in terms of using local image patches which also learns a semantic segmentation mask to select multiple keypoint locations from local patches belonging to an object and providing those inputs to the PnP algorithm. Regarding pose estimation using  synthetic datasets, Rozantsev \etal~\cite{Rozantsev2019BeyondSW} used a two-stream network trained on a synthetic and real dataset, and introduced loss functions that prevent corresponding weights of two streams from being too different from each other. Rad \etal~\cite{featuremapping} proposed a method that learns a feature mapping from real to synthetic datasets, and during inference transfers the features of real images to synthetic and infers pose using synthetic features. Some work has been done using a deep learning framework for Aqua2 vehicle detection that enabled visual servoing~\cite{ShkurtiIROS2017}.  Koreitem \etal~\cite{koreitem_oceans2018} used rendered images for pose estimation based visual tracking of Aqua2, and our approach  outperforms their approach in terms of 6D pose estimation accuracy.

Our work employs CycleGAN~\cite{CycleGAN2017}, a type of Generative Adversarial Network (GAN), to generate a synthetic dataset for training. GANs, introduced by Goodfellow \etal~\cite{gan}, are used to generate images through adversarial training where a generator attempts to produce realistic images to fool a discriminator which tries to distinguish if the image is real or generated. CycleGAN~\cite{CycleGAN2017} is used for unpaired image-to-image translation even in absence of corresponding real-generated image pair. The main idea of CycleGAN is that if an image is translated from one domain to another and translated back, the resulting image should resemble the original image.

\section{THE PROPOSED SYSTEM}\label{sec:proposed}

Figure \ref{fig:deepcl_pipeline} shows an overview of the proposed system. In the training process, UE4 renders a 3D model of Aqua2 with known 6D poses projected on top of underwater ocean images. The feature space between real underwater and rendered images is aligned by transferring the rendered images to target domains (swimming pool and ocean) using CycleGAN~\cite{CycleGAN2017}, an image-to-image translation network.

The next stage consists of a Convolutional Neural Network (CNN)  that predicts the 2D projections of the 8 corners of the object's (Aqua2) 3D model, similar to~\cite{bb8, single6dpose} and an object detection bounding box. Even though~\cite{bb8,single6dpose} divide an image into grid cells, they use global estimates of 2D keypoints for the object with the highest confidence value. In our approach, each grid cell inside the bounding box predicts the 2D projections of keypoints along with their confidences focusing on local regions belonging to the object. These predictions of all cells are then combined based on their corresponding confidence scores using RANSAC-based PnP during 6D pose estimation.

\subsection{Domain Adaptation}
\label{sec:domain}
We employ CycleGAN~\cite{CycleGAN2017} for unpaired image-to-image translation by learning functions to map the UE4 domain $R$ to the target domain $T$. We use generators G and F to transfer domains: $G:R\xrightarrow{} T$ and $F: T \xrightarrow{} R$. Discriminator, $D_R$, is designed to distinguish between rendered images in $R$, and augmented fake images $F(T)$. Discriminator, $D_T$, aims to separate target images in $T$ and augemented fake images $G(R)$. To improve image-to-image translation in CycleGAN, cycle consistency is maintained by ensuring the reconstructed images $F(G(R)) \approx    R$ in addition to the adversarial loss. To calculate adversarial loss, $G$ tries to generate $G(R)$, which is so similar to $T$ that can fool the discriminator $D_T$. The loss for $G$ and $D_T$ is:
\begin{equation}
    \label{loss:gan_loss}
    \begin{split}
    L_G(G,D_T,R,T) = E_{t\sim p_{\textrm{data}}(t)} [\log D_T (t)] + \\
                    E_{r\sim p_{\textrm{data}}(r)} [\log(1-D_T(G(r))] \hspace{5mm}    
    \end{split}
\end{equation}
where $t \sim p_{\textrm{data}}(t)$ and $r\sim p_{\textrm{data}}(r)$ denotes the data distribution in $T$ and $R$ respectively, and $E$ is the loss function, which is L1\hyp norm in our approach. Similarly we derive $L_R(F,D_R,T,R)$ following Eq. \ref{loss:gan_loss}. The cycle consistency loss $L_{\textrm{cyc}}$ is defined as: 
\begin{equation}
    \label{loss:gan_total}
    \begin{split}
        L_{\textrm{cyc}}(G,F) = E_{r\sim p_{\textrm{data}}(r)} [ || F(G(R)) - r ||_1 ] + \\
                        E_{t\sim p_{\textrm{data}}(t)}[ || G(F(t)) - t ||_1 ] 
    \end{split}
\end{equation}

In our proposed method, there are two target domains: swimming pool, $T_{sp}$ and an open-water ocean environment, $T_{m}$. Therefore, we train two instances of CycleGAN(two generators), $G_1:R \xrightarrow{}T_{sp}$ and $G_2:R\xrightarrow{}T_m$. Fig. \ref{fig:image_translation} shows the CycleGAN training overview along with synthetic data generation.

\begin{figure}[ht]
  \begin{center}
    \vspace{0.1in}
    \includegraphics[width=0.45\textwidth]{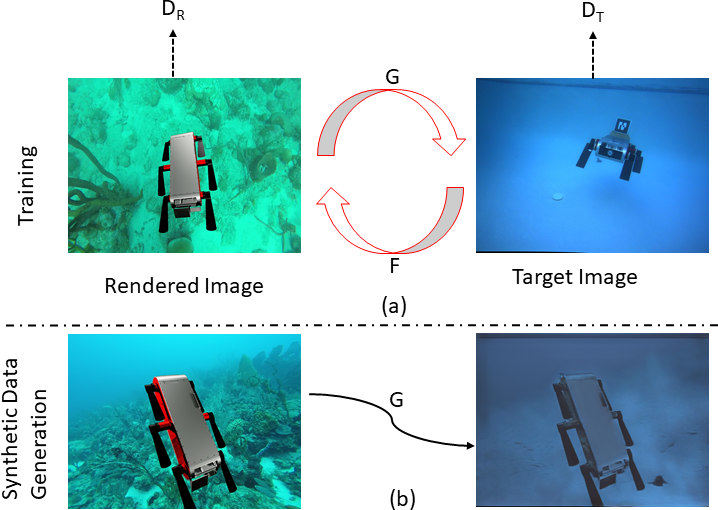}~
  \end{center}
  \caption{(a) CycleGAN learning process is shown. CycleGAN learns two mapping functions; $G:R\xrightarrow{}T$ and $F:T\xrightarrow{}R$ with two discriminators, $D_R$ and $D_T$. (b) Only using generator $G$, we perform image-to-image translation of rendered images $R$ to target images $T$.}
  \label{fig:image_translation}
\end{figure}
\invis{
 \begin{figure}[ht]
  \begin{center}
  \subfigure[]{\includegraphics[width=0.15\textwidth]{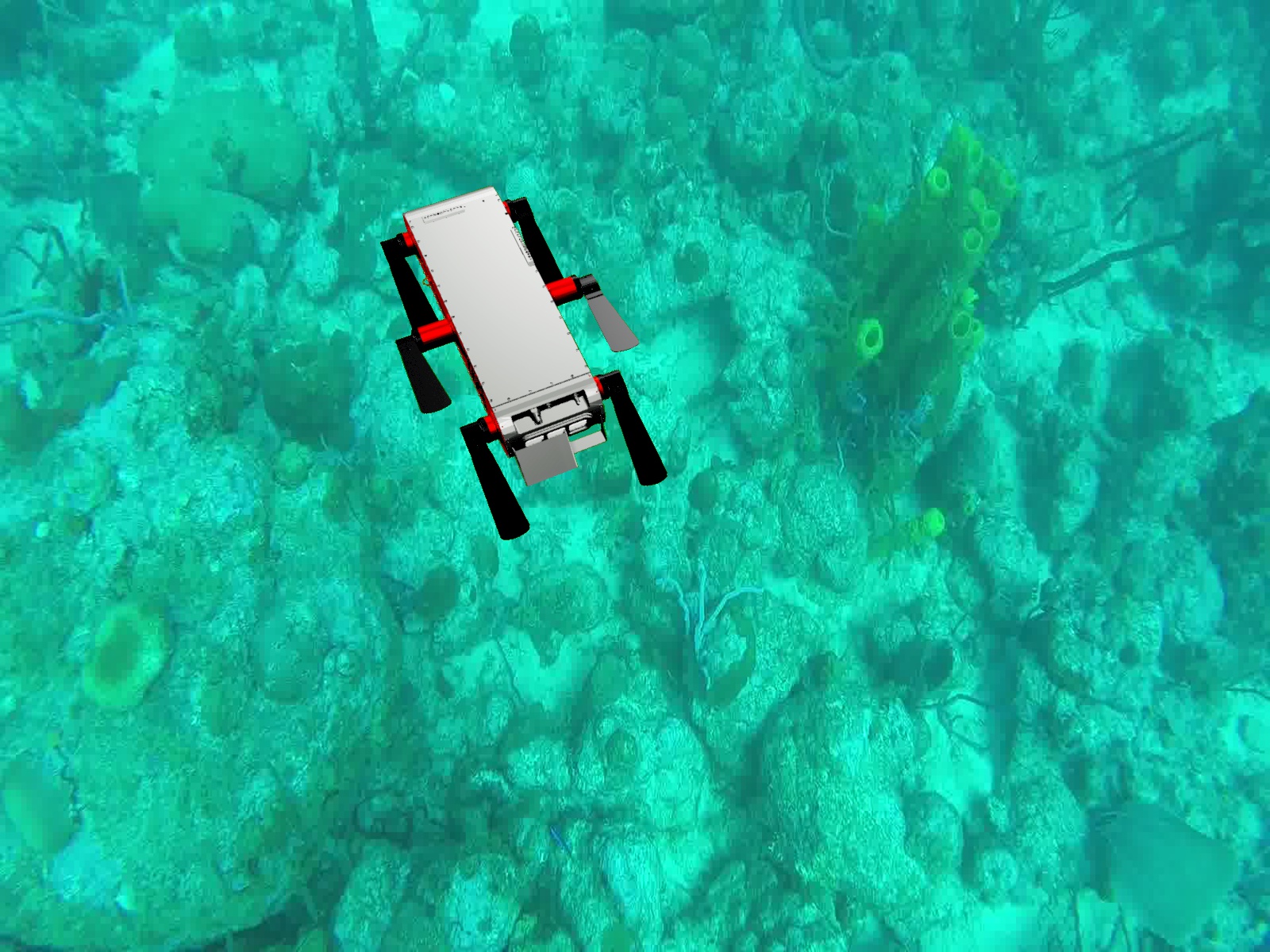}}~
  \subfigure[]{\includegraphics[width=0.15\textwidth]{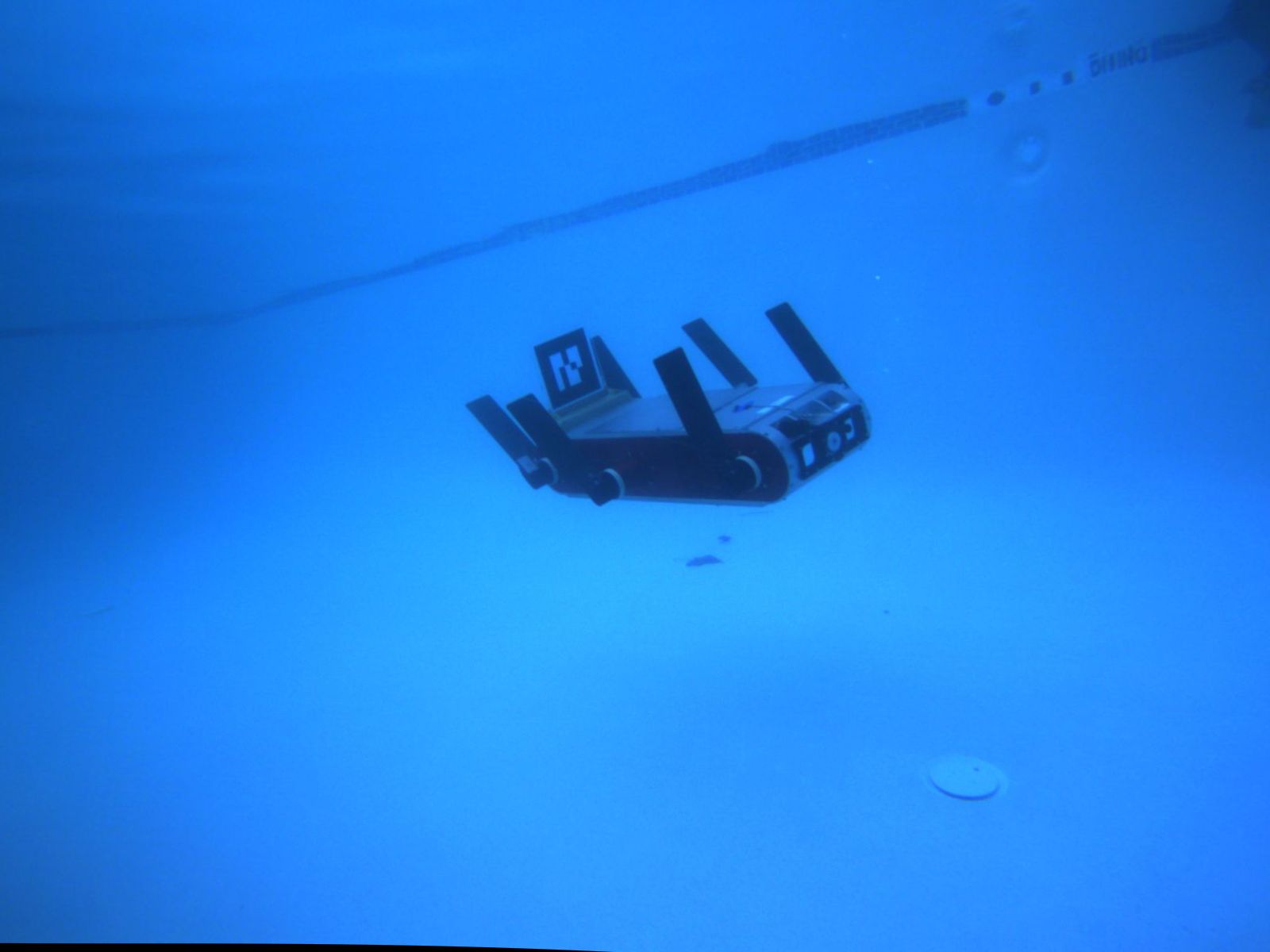}}~
  \subfigure[]{\includegraphics[width=0.15\textwidth]{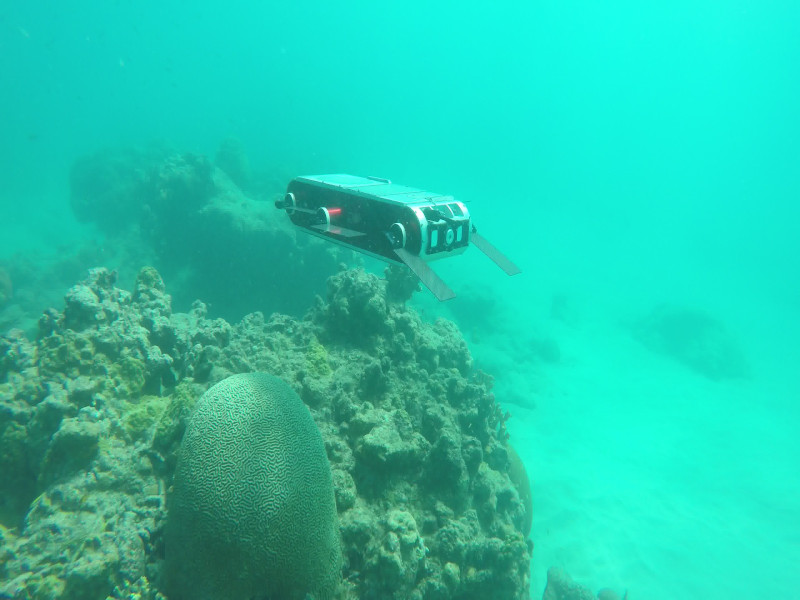}}~\\
    \subfigure[]{\includegraphics[width=0.15\textwidth]{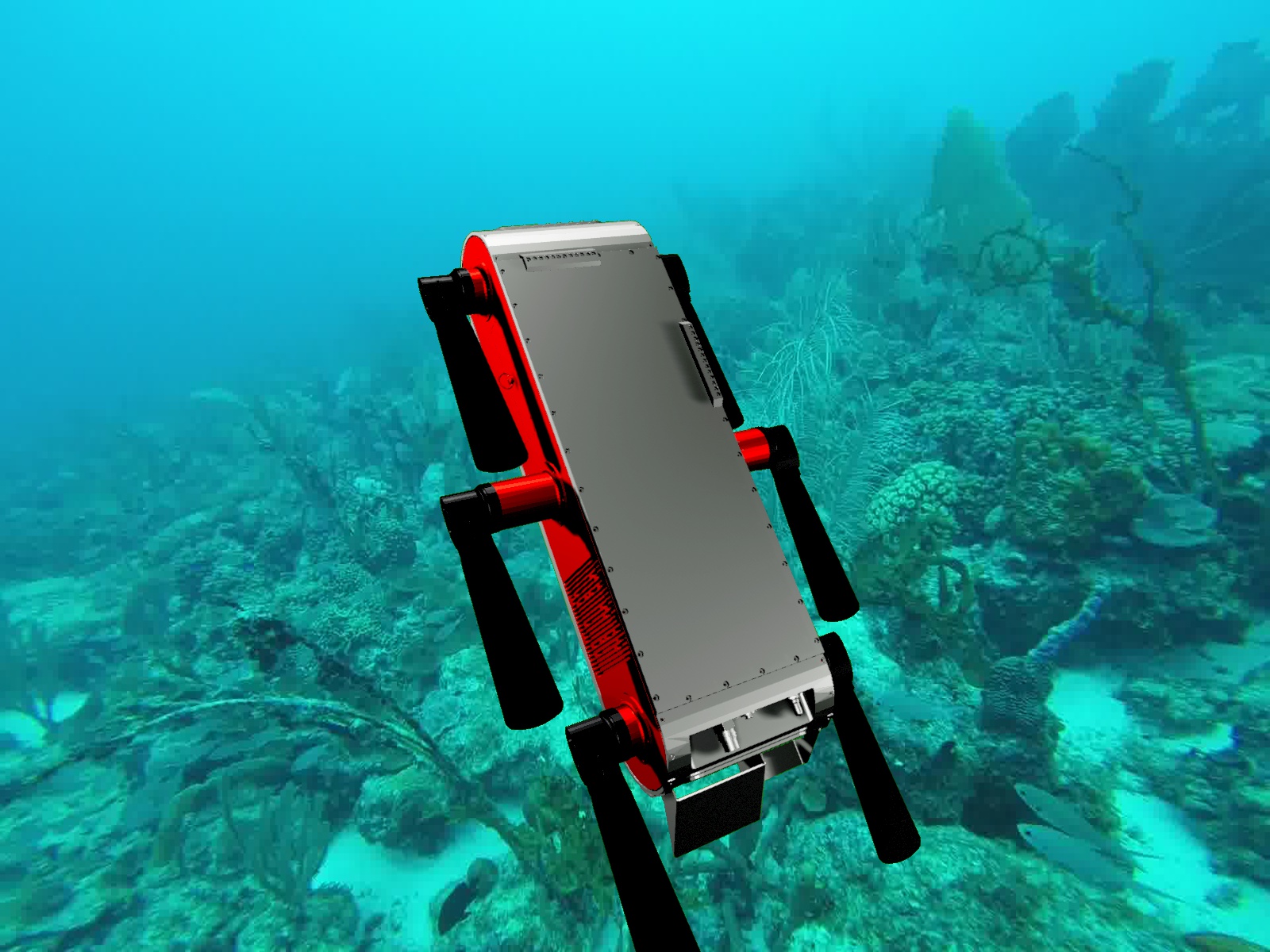}}~
    \subfigure[]{\includegraphics[width=0.15\textwidth]{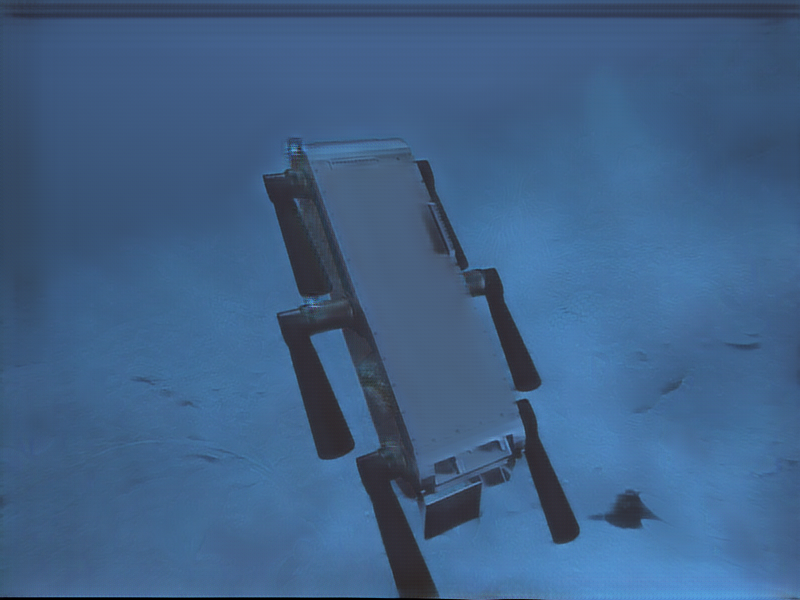}}~
    \subfigure[]{\includegraphics[width=0.15\textwidth]{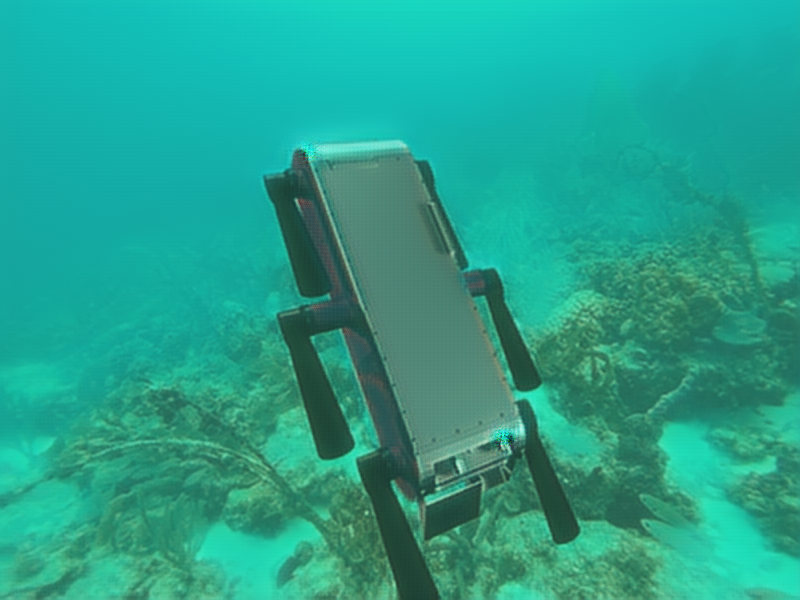}}~
  \end{center}
  \caption{(a) Rendered image by projecting Aqua2 model on top of underwater image (b) Real image of Aqua2 swimming in pool (c) Real image of Aqua2 shot using GoPro during ocean deployment (d) Another example of rendered image used to produce synthetic image (e) Synthetic image corresponding to (d) resembling Aqua2 in pool (b) using CycleGAN generator $G_1$. (f) Synthetic image corresponding  to (d) resembling Aqua2 in ocean environment using CycleGAN generator $G_2$}
  \label{fig:image_translation}
\end{figure}
}
\subsection{6D Pose Estimation}
The proposed network consists of an encoder, Darknet-53~\cite{yolov3}, and two decoders: Detection Decoder and Pose Regression Decoder. The detection decoder detects objects with bounding boxes, and the pose regression decoder regresses to 2D corner keypoints of the 3D object model. The decoders predict the output as a 3D tensor with a spatial resolution of $S \times S$ and a dimension of $D_{\textrm{det}}$ and $D_{\textrm{reg}}$, respectively. The spatial resolution controls the size of an image patch that can effectively vote for object detection and for the 2D keypoint locations. The feature vectors are predicted at three different spatial resolutions. The decoder stream detects features with multiple scales via upsampling and concatenation with a depth of final layer, $D_{\textrm{det}}$. The pose regression stream also has a similar architecture, but the final depth layer is maintained to be $D_{\textrm{reg}}$.  Predicting in multiple spatial resolutions with upsampling helps to obtain semantic information at multiple scales using fine-grained features from early on in the network.

\textbf{Object Detection Stream:} The object detection stream is similar to the detection stream of YOLOv3~\cite{yolov3} which predicts object bounding box. For each grid cell at offset $(c_x, c_y)$ from the top left corner of the image, the network predicts 4 coordinates for each bounding box $p_x, p_y, p_w, p_h$. Following~\cite{yolov3}, we use 9 anchor boxes obtained by k-means clustering on COCO dataset~\cite{coco} of size $(10 \times 13), (16 \times 30), (33 \times 23), (30 \times 61), (62 \times 45), (59 \times 119), (116 \times 90), (156 \times 198), (373 \times 326)$ divided among three scales. The width and height are predicted as the fraction of the anchor box priors $a_w, a_h$ and the actual bounding box values are obtained as
\begin{align}
    \label{eq:det}
    &b_x = \sigma(p_x) + c_x \nonumber \\
    &b_y = \sigma(p_y) + c_y \nonumber \\
    &b_w = a_we^{p_w} \nonumber \\
    &b_h = a_he^{p_h}  
\end{align}
where $\sigma$ represents the sigmoid function.
The sum of square of error between the ground truth $t_*$ and coordinate prediction $\hat{t_*}$ is used as the loss function. The ground truth values $t_*$ can be obtained by inverting equation \eq{eq:det}. The object detection stream also predicts the objectness score of each bounding box by calculating its intersection over union with anchor boxes and class prediction scores using independent logistic classifiers as in~\cite{yolov3}. The total object detection loss $L_{\textrm{det}}$ is the sum of coordinate prediction loss, objectness score loss, and class prediction loss. The total object detection loss was introduced by Redmon \etal~\cite{yolov1} to which we refer for a complete description.

 \begin{figure}[ht]
\vspace{0.1in}
\begin{center}
     \subfigure[]{\includegraphics[height=.15\textheight]{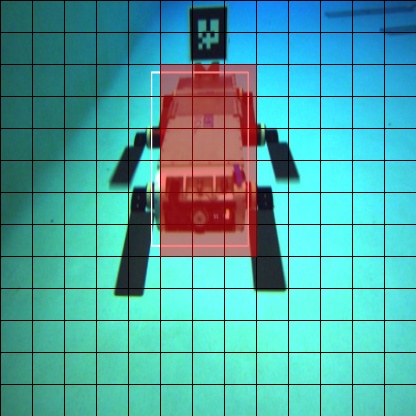}}~
     \subfigure[]{\includegraphics[height=.15\textheight]{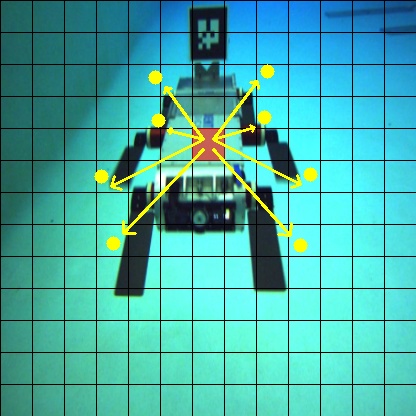}}~
 \end{center}
\vspace{-0.2in} \caption{ (a) The object detection stream predicts the bounding box and assigns each cell inside the box to the Aqua2 object. (b) The regression stream predicts the location of 8 bounding box corners as 2D keypoints from each grid cell.}
 \label{fig:streams}
\end{figure}

\textbf{Pose Regression Stream:} The pose regression stream predicts the location of the 2D projections of the predefined 3D keypoints associated with the 3D object model of Aqua2. We use 8 corner points of model bounding boxes as keypoints. The pose regression stream predicts a 3D tensor with size $S \times S  \times D_{\textrm{reg}}$. We predict the $(x,y)$ spatial locations for the 8 keypoint projections along with their confidence values, $D_{\textrm{reg}}=3\times8$.

We do not predict the 2D coordinates of the 2D keypoints directly. Rather, we predict the offset of each keypoint from the corresponding grid cell as in \fig{fig:streams}(b) in the following way: Let $c$ be the position of grid cell from top left image corner. For the $i^{th}$ keypoint, we predict the offset $f_i(c)$ from grid cell, so that the actual location in image coordinates becomes $c + f_i(c)$, which should be close to the ground truth 2D locations $g_i$. The residual is calculated as 
\begin{equation}
    \Delta_i(c) = c + f_i(c) - g_i
    \label{eq:residual}
\end{equation}
and we define offset loss function, $L_{\textrm{off}}$, for spatial residual:
\begin{equation}
    \label{eq:loss_pos}
    L_{\textrm{off}} = \sum_{c \epsilon B} \sum_{i=1}^{8} \left \lvert \lvert \Delta_i(c) \right \rvert \rvert_1
\end{equation}
where $B$ consists of grid cells that fall inside the object bounding box and $\left \lvert \lvert . \right \rvert \rvert_1 $ represents L1\hyp norm loss function, which is less susceptible to outliers than L2 loss. Only using grid cells falling inside the object bounding box for 2D keypoint predictions focuses on image regions that truly belong to the object.

Apart from the 2D keypoint locations, the pose regression stream also calculates the confidence value $v_i(c)$ for each predicted point, which is obtained through the sigmoid function on the network output. The confidence value should be representative of the distance between the predicted keypoint and ground truth values. A sharp exponential function of the 2D euclidean distance between prediction and ground truth is used as confidence. The confidence loss is calculated as 
\begin{equation}
    \label{eq:loss_conf}
    L_{\textrm{conf}} = \sum_{c \epsilon B} \sum_{i=1}^{8} \left \lvert \lvert v_i(c) - exp(-\alpha \left \lvert \lvert \Delta_i(c) \right \rvert \rvert_2) \right \rvert \rvert_1
\end{equation}
where $\left \lvert \lvert . \right \rvert \rvert_2 $ denotes euclidean distance or L2 loss and parameter $\alpha$ defines the sharpness of the exponential function. The pose regression loss of \eq{eqn:loss} takes up the form

\begin{equation}
    \label{eqn:pose_loss}
    L_{\textrm{reg}} = \lambda_{\textrm{off}}L_{\textrm{off}} + \lambda_{\textrm{conf}}L_{\textrm{conf}}
\end{equation}

For numerical stability, we down-weight the confidence loss for cells that do not contain objects by setting $\lambda_{\textrm{conf}}$ to 0.1, as suggested in~\cite{yolov1}. For the cells that include the object, $\lambda_{\textrm{conf}}$ is set to 5.0 and $\lambda_{\textrm{off}}$ to 1. Therefore, the total loss of the network is:
\begin{equation}
    \label{eqn:loss}
    L = L_{\textrm{det}} + L_{\textrm{reg}}
\end{equation}

 \begin{figure}[ht]
 \begin{center}
     \subfigure[]{\includegraphics[height=.15\textheight]{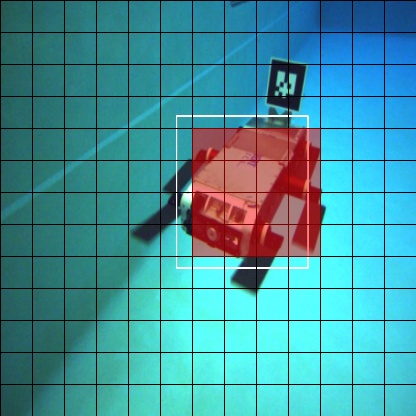}}~
     \subfigure[]{\includegraphics[height=.15\textheight]{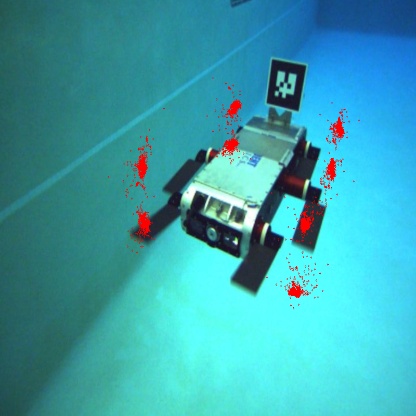}}~\\
    \subfigure[]{\includegraphics[height=.15\textheight]{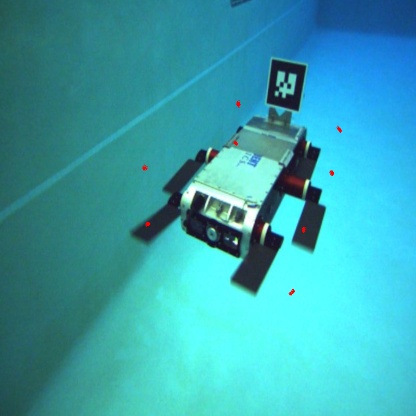}}~
    \subfigure[]{\includegraphics[height=.15\textheight]{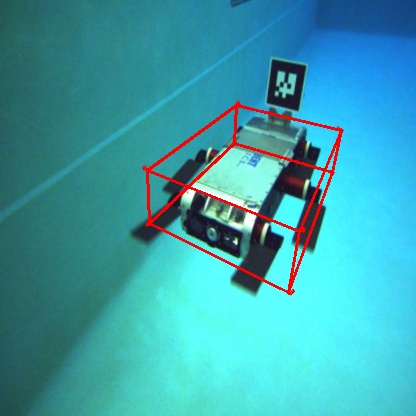}}~
 \end{center}
 \vspace{-0.2in}\caption{ Inference strategy for combining pose candidates. (a) Grid cells inside the detection box belonging to Aqua2 object overlaid on the image. (b) Each grid predicts 2D locations for corresponding 3D keypoints shown as red dots. (c) For each keypoints, 12 best candidates are selected based on the confidence scores. (d) Using $12\times8=96$ 2D-to-3D correspondence pairs and running RANSAC\hyp based PnP algorithm yield accurate pose estimate as shown by the overlaid bounding box. }
 \label{fig:inference}
\end{figure}

\subsection{Pose Refinement}
During inference, the object detection stream of our network predicts the coordinate locations of the bounding boxes with their confidences and the class probabilities for each grid cell. Then, the class-specific confidence score is estimated for the object by multiplying the class probability and confidence score. To select the best bounding box, we use non-max suppression~\cite{nonmaxsupression} with an IOU threshold of 0.4 and a class-specific confidence score threshold of 0.3.

Simultaneously, the pose regression stream produces the projected 2D locations of the object's 3D bounding box, along with their confidence scores for each grid cell, as shown in \fig{fig:inference}-b. The 2D keypoint predictions for grid cells that fall outside of the bounding box (\fig{fig:inference}-a) from the object detection stream are filtered out. In an ideal case, the remaining 2D keypoints should cluster around the object center. 2D keypoints that do not belong to a cluster are removed using a pixel distance threshold of 0.3 times image width. The keypoints with confidence scores less than 0.5 are also filtered out. To balance the trade-off between computation time and accuracy, we empirically found that using the 12 most confident 2D predictions for each 3D keypoint (\fig{fig:inference}-c) produces an acceptable pose estimate after RANSAC\hyp based PnP~\cite{epnp}. Hence, we employ RANSAC\hyp based PnP~\cite{epnp} on $12\times8=96$ 2D\hyp to\hyp 3D correspondence pairs between the image keypoints and the object's 3D model to obtain a robust pose estimate, as shown in \fig{fig:inference}-d.

\invis{OLD ONE:At the same time, the pose regression steam produces the projected 2D locations of the object's 3D bounding box, along with their confidence scores for each grid cell. The 2D keypoint predictions for grid cells that fall outside the bounding box from the object detection stream are filtered out. In an ideal case, these 2D keypoints should cluster around the object center. Therefore, 2D keypoints that do not belong to a cluster are removed using a pixel distance threshold of 0.3 times image width. The keypoints with confidence scores less than 0.5 are also filtered out. We employ RANSAC\hyp based PnP~\cite{epnp} on the remaining predictions to obtain 2D\hyp to\hyp 3D correspondence between the image keypoints and the object's 3D model. However, this will make the predictions slow due to the optimization process. Empirically, we found using the 12 most confident 2D predictions for each 3D keypoint produces an acceptable pose estimate with a good trade\hyp off between computation time and accuracy, as shown in \fig{fig:inference}.}

\subsection{Implementation Details}
To create the synthetic dataset, we train the CycleGAN following the training procedure of~\cite{CycleGAN2017}. We let the training continue until it generated acceptable reconstruction. Once CycleGAN can reasonably reconstruct for the target domain, we use the model weights of that epoch to translate all rendered images to synthetic images. Then, the synthetic images are scaled to $416\times416$ resolution maintaining the aspect ratio  by padding zeros for training. During inference, no augmentation is required, and the real images are directly fed to the network.

The CNN is trained for 125 epochs on the synthetic dataset, and the first 3 epochs are part of a warmup phase, where the learning rate gradually increases from 0 to 1e-4. We utilized the SGD optimizer with a momentum of 0.9 and a piecewise decay to decrease the learning rate to 3e-5 and 1e-5 at 60 and 100 epochs, respectively. To avoid overfitting, minibatches of size 8 were produced by applying data augmentation techniques, including randomly changing hue, saturation, and exposure of the image up to a factor of 1.5. In addition, images were randomly scaled, and affine transformed by up to 25\% of the original image size.

\section{EXPERIMENTS}\label{sec:experiments}
This section describes first the datasets used, and then results of the inference with the real Aqua2 robot swimming in both a pool and the open ocean at Barbados.

\subsection{Datasets Description}
\label{desc:dataset}

\textbf{ Training - Rendered/Synthetic Dataset:} contains images obtained by rendering an Aqua2 robot swimming with flipper motion using UE4 and overlaying the resulted 3D model over random underwater images. Rather than just overlaying the 3D model of Aqua2, we simulate the flipper motion to generate images with the flippers in various realistic positions. This flipper motion makes the neural network independent of the flipper position. The synthetic dataset is obtained by the image-to-image translation network based on CycleGAN described in \ref{sec:domain} to create photo-realistic images. The rendered dataset contains 37K images with random depth between  \SI{0.75}{\m} and \SI{3.0}{\m} and orientations ranging from $-50$ to $50$ degrees for roll, $-70$ to $70$ degrees for pitch, and $-90$ to $90$ degrees for yaw. 

\begin{figure*}[ht]
\vspace{0.1in} 
\begin{center}
     \subfigure[]{\includegraphics[height=.12\textheight]{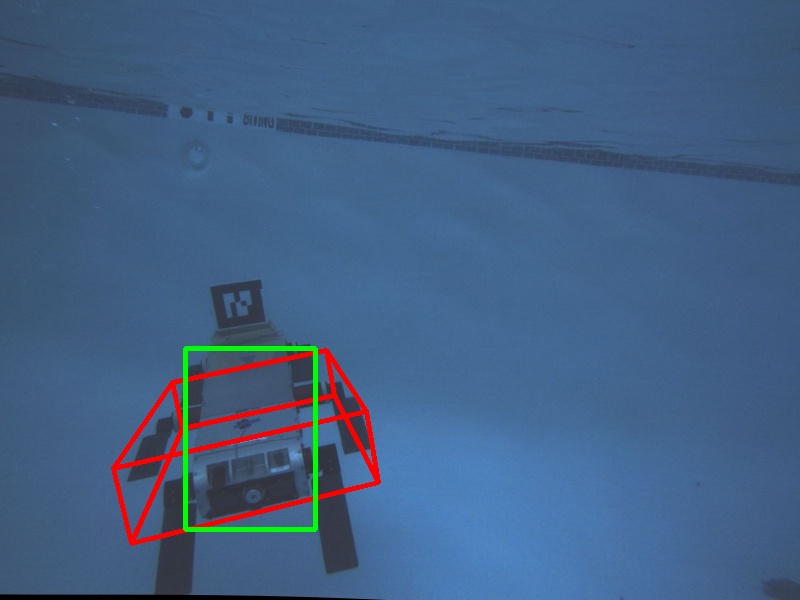}}~
     \subfigure[]{\includegraphics[height=.12\textheight]{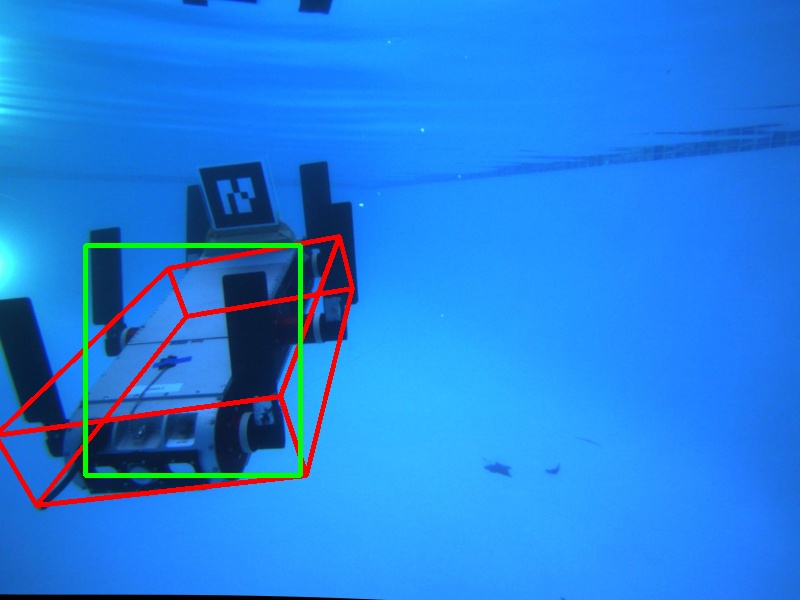}}~
    \subfigure[]{\includegraphics[height=.12\textheight]{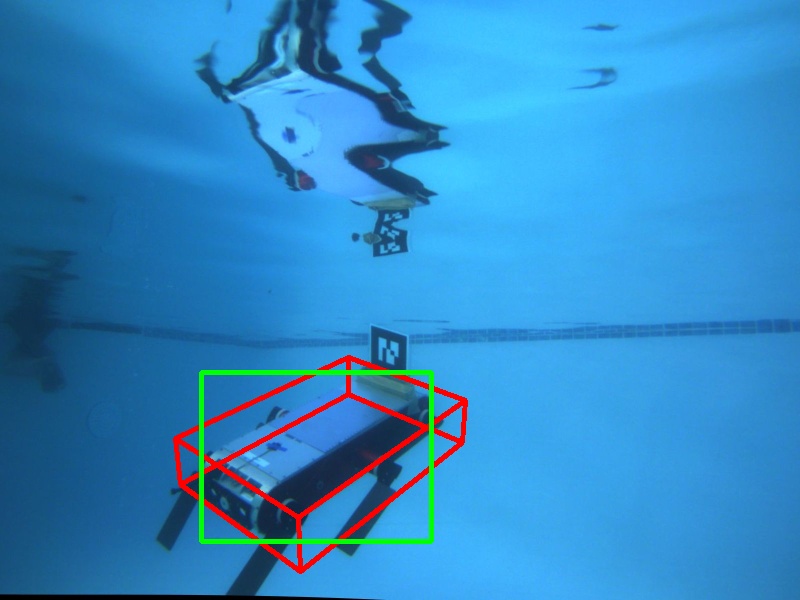}}~
    \subfigure[]{\includegraphics[height=.12\textheight]{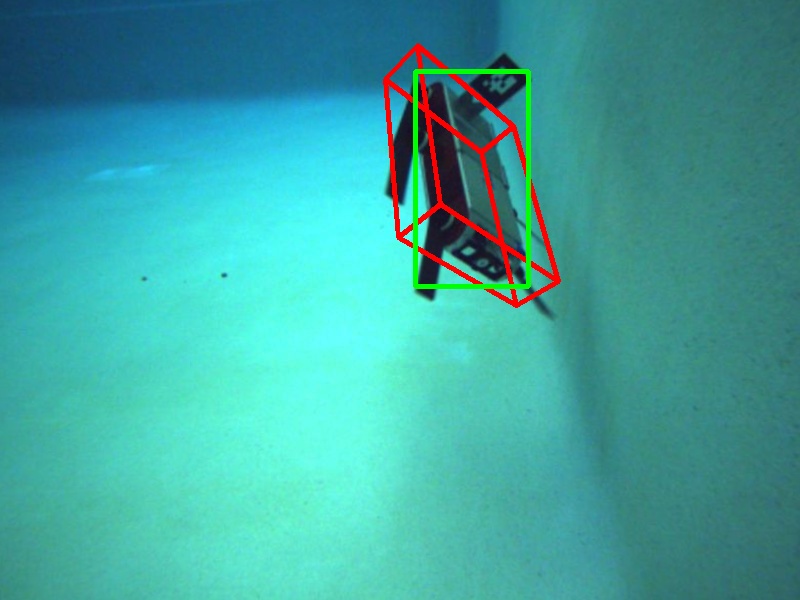}}~ \\
    \subfigure[]{\includegraphics[height=.09\textheight]{figures/gopro_0.png}}~
     \subfigure[]{\includegraphics[height=.09\textheight]{figures/gopro_1.png}}~
    \subfigure[]{\includegraphics[height=.09\textheight]{figures/gopro_2.png}}~
    \subfigure[]{\includegraphics[height=.09\textheight]{figures/gopro_3.png}}~\\
    \subfigure[]{\includegraphics[height=.12\textheight]{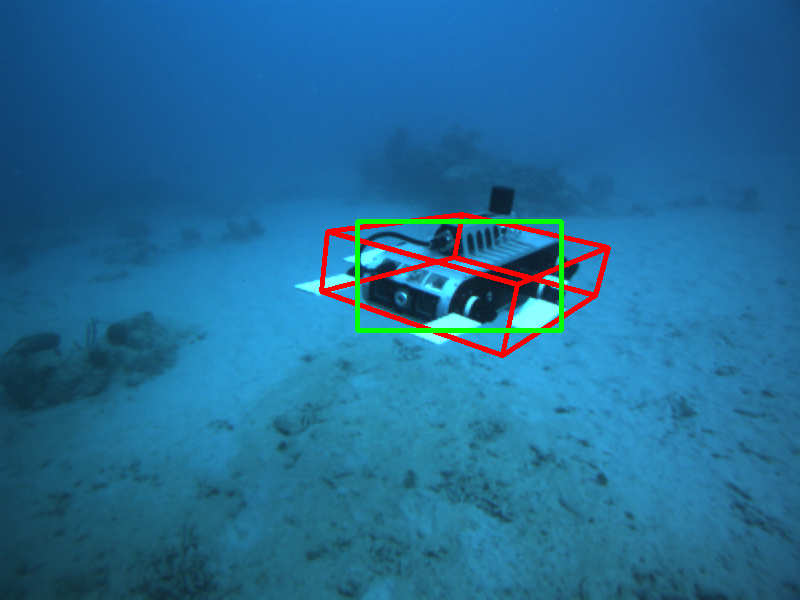}}~
     \subfigure[]{\includegraphics[height=.12\textheight]{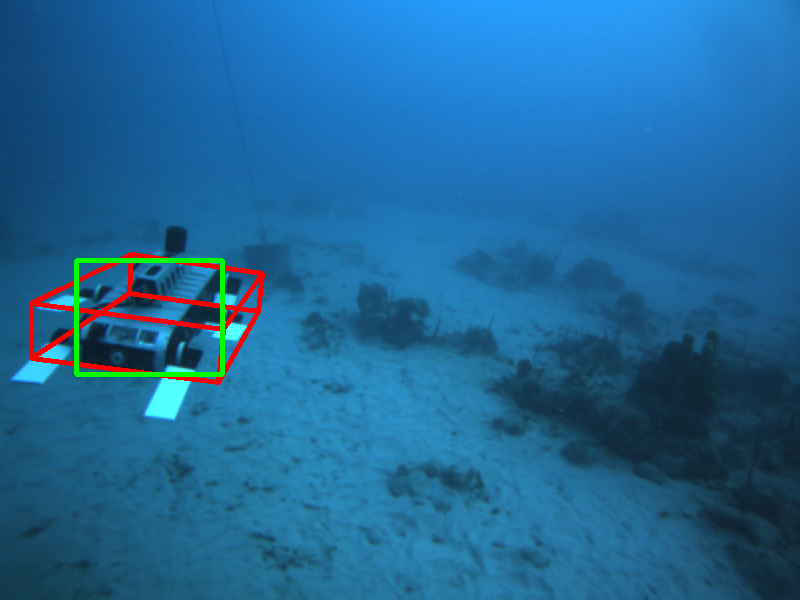}}~
    \subfigure[]{\includegraphics[height=.12\textheight]{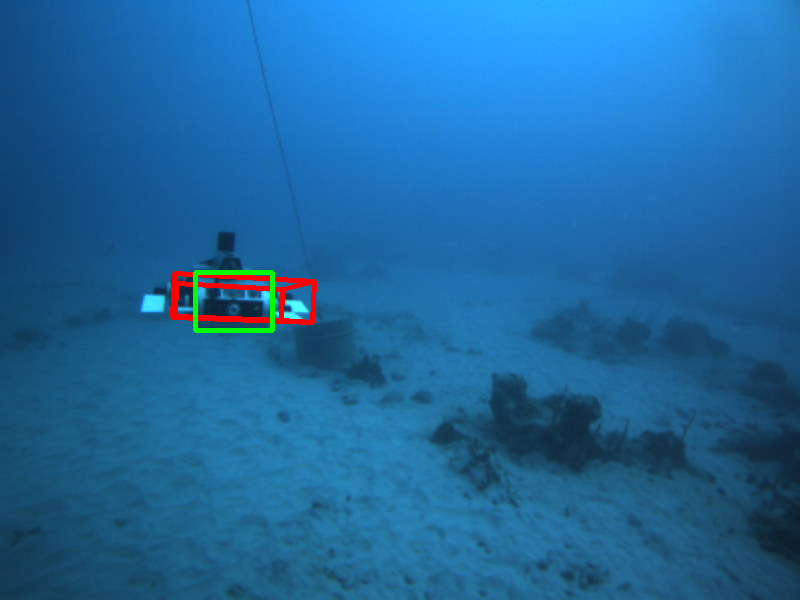}}~
    \subfigure[]{\includegraphics[height=.12\textheight]{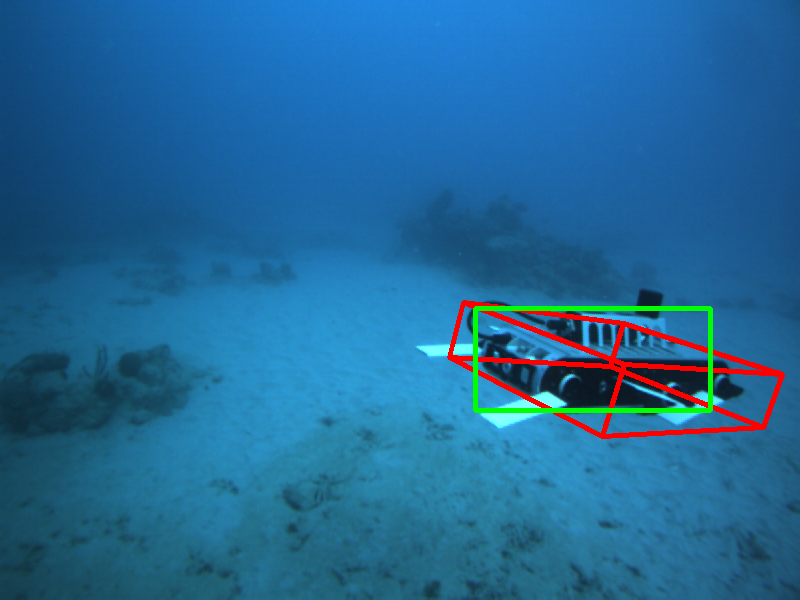}}~ 
    
 \end{center}
 \vspace{-0.2in}\caption{Sample detections from the different datasets. Green square is the 2D detection box, while the red wireframe is the projection of the 3D bounding box of the robot. Top row: Pool dataset, observed Aqua2 vehicle carries a AR tag to generate ground truth estimates; observing robot is another Aqua2. Middle row: GoPro footage during deployments in Barbados in January 2020, observed robot has no additional components, the observing camera is a GoPro 7 camera. Bottom row: Barbados 2017 dataset~\cite{koreitem_oceans2018}, observed robot is equipped with a Ultra-Short Baseline (USBL) modem, the observing robot is an  Aqua2 vehicle.}
 \label{fig:all_results}
\end{figure*}

\textbf{Testing - Pool Dataset:} To generate our pool dataset, we deployed two robots: one robot observing the other with a vision-based 2D fiducial marker (AR tag\footnote{\url{http://wiki.ros.org/ar_track_alvar}}) mounted on the top used to estimate ground truth during two pool trials in indoor and outdoor pools, as shown in Fig. \ref{fig:all_results}(a-d). Approximately, 11K images were collected with estimated localization, provided from the pose detection of the AR tag and the relative transformation of the mounted tag to the real robot. The dataset contains images with a distance between two Aquas ranging between 0.5m to  3.5m.

\textbf{Testing - Barbados 2017 Dataset:} The Barbados 2017 Dataset consists of 188 real images collected during underwater field trials off the west coast of Barbados used in~\cite{koreitem_oceans2018}, see \fig{fig:all_results}(i-l). The images are captured from an Aqua2 robot's onboard camera. 6D pose of the robot in each of these images is obtained using a custom-built annotator, which allows the user to mark keypoints on the robot assigned from the CAD model. The annotator then iteratively fits a wireframe to the robot using its known dimensions.

\textbf{Testing - Barbados GoPro Dataset:} We collected images underwater in Barbados of an Aqua2 robot  swimming over coral reefs using a GoPro camera, which differ significantly from the images collected using another Aqua2 in terms of hue, image size, and aspect ratio (see \fig{fig:all_results}(e-h)). Given that ground truth is unavailable for these images, this dataset was only used to evaluate the proposed method qualitatively.
\invis{
 \begin{figure}[ht]
 \begin{center}
     \subfigure{\includegraphics[height=.10\textheight]{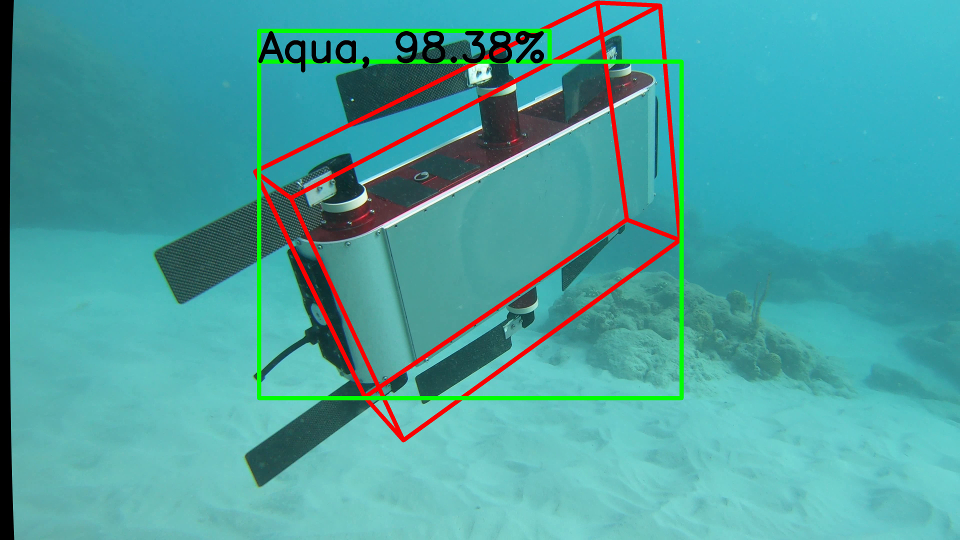}}~
     \subfigure{\includegraphics[height=.10\textheight]{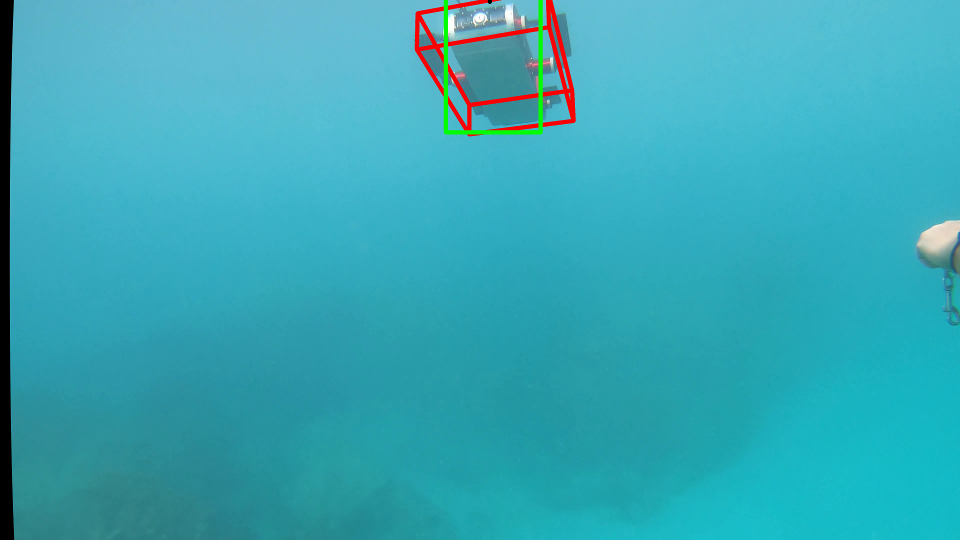}}~\\
    \subfigure{\includegraphics[height=.10\textheight]{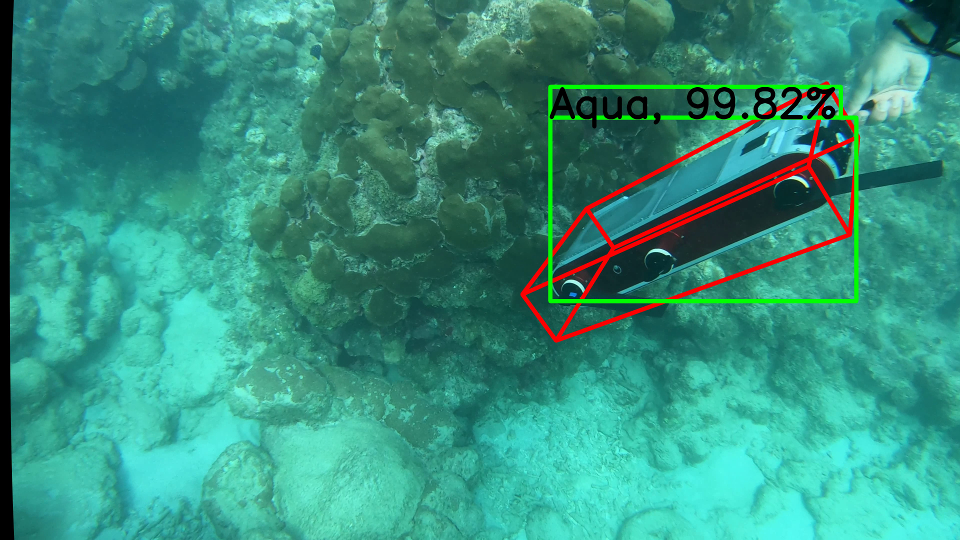}}~
    \subfigure{\includegraphics[height=.10\textheight]{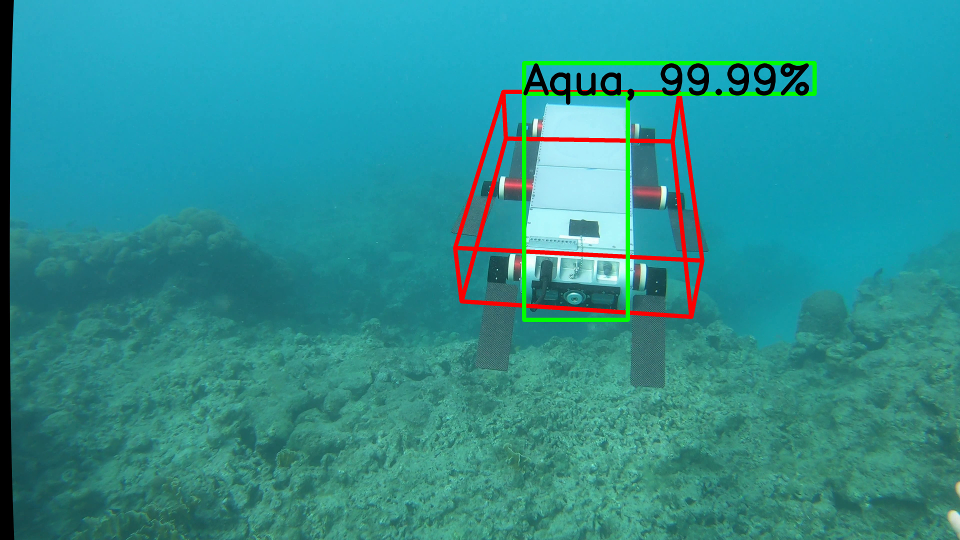}}~
 \end{center}
 \caption{Qualitative evaluation on Barbados GoPro Dataset}
 \label{fig:qual_gopro}
\end{figure}
}

\begin{table}
\vspace{0.1in}    \centering
    \resizebox{0.485\textwidth}{!}{
    \begin{tabular}[t]{@{} l|cccc|c}
    \toprule
    & \bf{\makecell{Translation \\ Error}} & \bf{\makecell{Orientation \\ Error}} & \bf{\makecell{REP-10px \\ Accuracy}} & \bf{\makecell{ADD-0.1d \\ Accuracy}}  & \bf{FPS}\\
    \midrule
    
    Tekin \etal~\cite{single6dpose} & 0.278m & 18.87\degree & 9.33\% & 23.39\% & \bf{54}\\
    
    PVNet ~\cite{peng2019pvnet} & 0.486m & 24.55\degree & 23.22\% & 43.09\% & 37\\
    
    DeepURL & \bf{0.068m} & \bf{6.77\degree} & \bf{25.22\%} & \bf{57.16\%} & 40 \\
    
    \bottomrule
    \end{tabular}}
    \caption{Translation  and  Orientation  errors (the lower the better) along with REP-10px, ADD-0.1d accuracy (the higher the better) and runtime comparison for the pool dataset}
    \label{tab:pose_estimation_error}
\end{table}
\subsection{Evaluation Metrics}
To evaluate the pose estimation capability of the proposed system, we calculated the mean translation error as the Euclidean distance between the predicted and the ground truth translation. Let $(\textrm{Rot},\textrm{trans})$ and $(\hat{\textrm{Rot}},\hat{\textrm{trans}})$ be the ground truth and predicted rotation matrices and translation, respectively. For individual angle errors in terms of yaw, roll, and pitch, we decomposed the rotation matrices ${Rot}$ and $\hat{Rot}$ into Euler angles and calculated their absolute difference. The total orientation error is represented as \eq{eqn:rotation_error}, where $tr$ represents the trace of the matrix and the orientation error is in the range of $[0,\pi]$.
\begin{equation}
    \label{eqn:rotation_error}
   \phi ({\textrm{Rot},\hat{\textrm{Rot}}}) = \arccos{\frac{tr({{\textrm{Rot}}^T}\hat{\textrm{Rot}}) - 1}{2}}
\end{equation}
To evaluate the pose accuracy, we use standard metrics - namely- 2D reprojection error~\cite{Brachmann_2016_CVPR} and the average 3D distance of the model vertices, referred to as ADD metric~\cite{xiang2017posecnn, bb8}. In the case of reprojection error, we consider the pose estimate as correct if the average distance between 2D projections of 3D model points obtained using predicted and ground-truth poses is below a 10 pixels threshold, referred to as REP-10px. Generally, a 5-pixel threshold is employed, but we consider a threshold of 10 pixels to account for uncertainties in ground truth due to the AR tag-based pose estimation. The ADD metric takes pose estimate as correct if the mean distance between the coordinates of the 3D model vertices transformed by estimated and ground truth pose fall below 10\% of the model diameter, referred to as ADD-0.1d.  We also report the inference time of the algorithm in terms of frames per second (FPS) on an RTX 2080 GPU.

\subsection{Experimental Results}

\textbf{Evaluation on the Pool Dataset:} We compare our method with the state-of-the-art method of Tekin \etal~\cite{single6dpose} and PVNet~\cite{peng2019pvnet} trained on a synthetic dataset and tested on real pool dataset. Translation and rotation errors along with REP-10px and ADD-0.1d accuracy for the pool dataset are presented in \tab{tab:pose_estimation_error}, as well as the runtime comparisons on Nvidia RTX 2080. DeepURL outperforms both the method of Tekin \etal ~\cite{single6dpose} and PVNet~\cite{peng2019pvnet} in terms of rotation and translation errors along with REP-10px and ADD-0.1d accuracy. Moreover, the runtime performance is realtime, outperforming PVNet~\cite{peng2019pvnet} and only slightly slower than that of Tekin \etal~\cite{single6dpose}. The improved performance comes from two factors: 1) the use of a better detection pipeline and 2) bounding box based keypoint sampling introduced in this paper. Whereas Tekin \etal~\cite{ssd_6d} only used the keypoints with the highest confidence, our bounding box based keypoint sampling allows the selection of more appropriate keypoints using RANSAC-based PnP. PVNet~\cite{peng2019pvnet}, compared to DeepURL, performed slightly inferior on REP-10px and ADD-0.1d metrics and produced significantly higher translation and orientation errors. \invis{PVNet~\cite{peng2019pvnet} uses a segmentation mask to select pixels that are used to predict the direction of the unit vector from every selected pixel to 2D keypoint. An offset in  segmentation mask prediction causes the keypoint prediction to shift as every pixel inside the segmentation mask votes for 2D keypoints.}

\invis{DeepURL achieves 0.068m against 0.278m for Tekin \etal~\cite{ssd_6d}. DeepURL also estimates the orientation within $\sim 7\degree$ error as opposed to $\sim 19\degree$ in Tekin \etal~\cite{ssd_6d}. DeepURL also outperforms Tekin \etal~\cite{ssd_6d} significantly on REP-10px and ADD-0.1d accuracy. The improved performance comes from two factors, specifically: the use of better detection pipeline and bounding box based keypoint sampling introduced in this paper. The method is highly dependent on the accurate and precise detection of the Aqua2. As known in the literature, YOLOv3 performance significantly better YOLOv1 that is used in Tekin \etal~\cite{ssd_6d}. Again in our experiments, we saw that Tekin \etal~\cite{ssd_6d} cannot detect Aqua2 when it was further away from the camera. Secondly, the bounding box based keypoint sampling also allowed the selection of more appropriate keypoints using RANSAC-based PnP, whereas Tekin \etal~\cite{ssd_6d} only used the keypoints with the highest confidence.}

\begin{figure}[ht]
\begin{center}
\subfigure[]{\includegraphics[width=.24\textwidth]{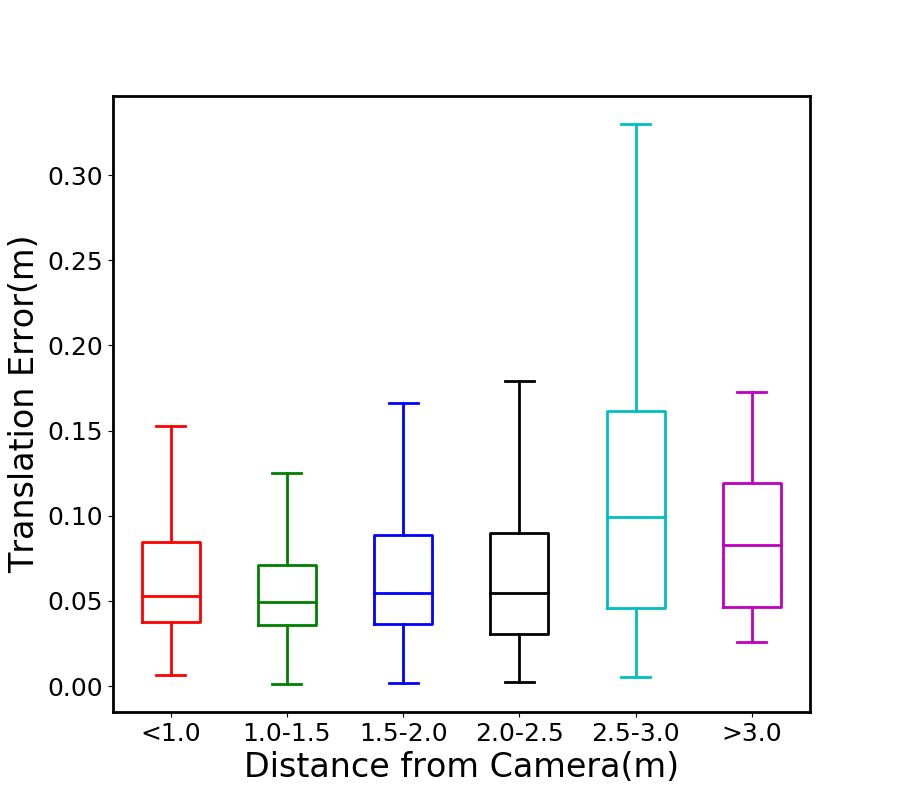}}\vspace{0.1em}%
 \subfigure[]{\includegraphics[width=.24\textwidth]{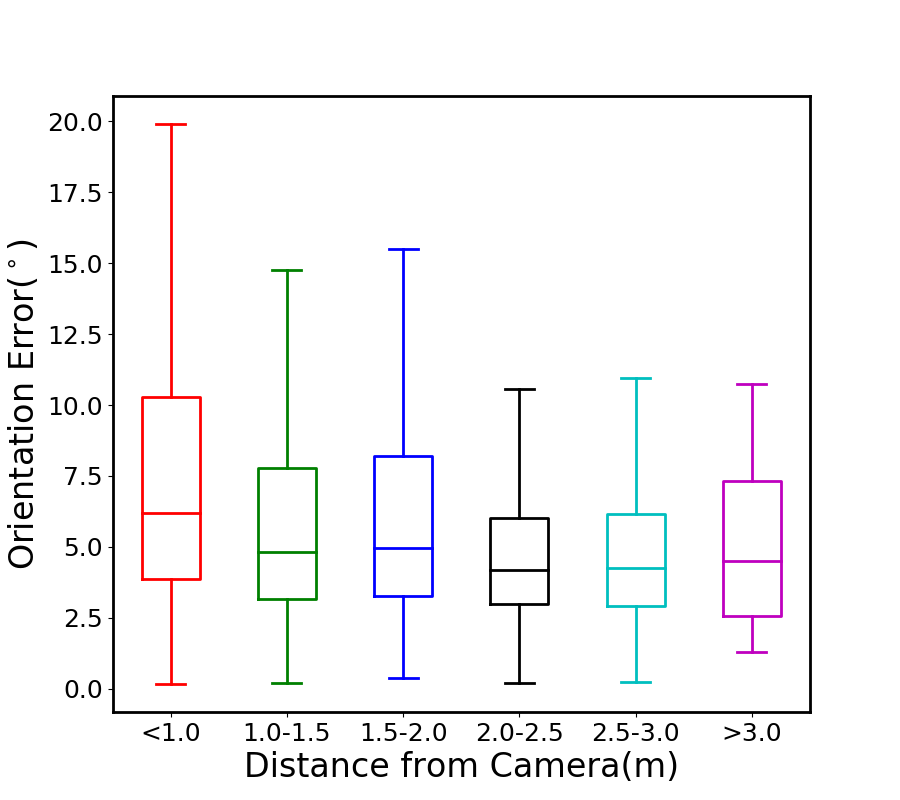}}~
\end{center}
\vspace{-0.2in}
\caption{Boxplot summarizing the error statistic of (a) translation and (b) orientation in Pool Dataset with respect to variable distance from camera.}
\label{fig:error_dist}
\end{figure}

Figure \ref{fig:error_dist} shows the translation and orientation error statistics of an Aqua2 robot in the pool dataset. It is evident that the proposed method performs well across  all distances from camera (0.5m-3.5m). Interestingly, at very close distance, the method experiences higher orientation error due to the 2D keypoints of Aqua2 not being precisely selected by the pose regression decoder. 

\textbf{Evaluation on the Barbados 2017 Dataset:} We report the performance of our system on Real Barbados 2017 dataset in terms of translation and rotation errors as shown in \tab{tab:mcgill_eval}. DeepURL performs significantly better on translation error and orientation error compared to the method of Koreitem \etal~\cite{koreitem_oceans2018}.

\begin{table}[b]
    \centering
    \resizebox{0.485\textwidth}{!}{
    \begin{tabular}[t]{@{} l|ccccc}

    \toprule
    & \bf{\makecell{Translation \\ Error}} &\bf{\makecell{Orientation \\ Error}} & \bf{\makecell{Roll \\ Error}} & \bf{\makecell{Pitch \\ Error}} & \bf{\makecell{ Yaw \\ Error}}  \\
    \midrule
    
    Koreitem \etal~\cite{koreitem_oceans2018} & 0.72m & 17.59\degree & 11.87\degree & 4.59\degree & 12.11\degree\\
    
    DeepURL & \bf{0.31m} & \bf{11.98\degree} & \bf{9.64\degree} & \bf{3.30\degree} & \bf{5.43\degree} \\
    \bottomrule
    \end{tabular}}
    \caption{Translation and Rotation errors for the Barbados 2017 dataset~\cite{koreitem_oceans2018}}\
    \label{tab:mcgill_eval}
\end{table}

\textbf{Impact of Domain Adaptation:} To understand the efficacy of using CycleGAN based domain adaptation, we trained DeepURL only on rendered images. Even though the network performed well on the validation set consisting of rendered images only, without training on synthetic dataset it was not able to generalize to real-world pool images. The intuition is that the real-world underwater images differ significantly from rendered images in terms of texture, color and appearance. Thus, creating image sets with different appearance and texture helps extensively in the training process by reducing over-fitting and increasing generalization.

\textbf{Failing Scenarios:} Predictions of pose estimate might be wrong either when the detection stream fails to predict the object detection box, therefore, there is not enough points for the RANSAC\hyp based PnP algorithm (at least six 2D-to-3D correspondences are required), or PnP did not converge. These detection failure scenarios are inherent in YOLOv3 architecture. The system may also fail for position or orientation not introduced in the training scenarios, such as translation beyond 3.5m or orientation beyond the rendered range described in Section \ref{desc:dataset}.

\section{CONCLUSION}
\label{sec:conclusions}

In this work, we presented a system for 6D pose estimation of an autonomous underwater vehicle for relative localization underwater. The system learns to predict the 6D pose without the need for any real ground truth, which enables pose estimation in an environment where ground truth is difficult to acquire. We also present a detection bounding box based keypoint sampling strategy that is more robust to related  work~\cite{peng2019pvnet,single6dpose} which leads to a better estimate of the pose of the observed robot, up to an order of magnitude is some cases; see Table \ref{tab:pose_estimation_error}.

Currently, the proposed network is being ported to an Intel Neural Compute Stick 2 (Intel NCS2)\footnote{\url{https://software.intel.com/en-us/neural-compute-stick}} and an NVidia Jetson TX2 Module\footnote{\url{https://developer.nvidia.com/embedded/jetson-tx2}} in order to deploy on an Aqua2 or a BlueROV2 vehicle. The above two platforms were selected based on their performance~\cite{ModasshirCRV2018} and compatibility with the proposed vehicles. Furthermore, the DeepURL framework will be integrated with the proprioceptive sensors of each robot (IMU and depth) and either the USBL positioning of the observer or the Visual\hyp Inertial estimator~\cite{RahmanIROS2019a} to recover the pose of both robots in a global frame of reference. 

\invis{The proposed method will enable robust operations underwater for the Aqua2 robots and is expected to be applicable to other underwater vehicles.}

%
\bibliographystyle{IEEEtran}
\bibliography{IEEEabrv,refs}

\begin{thebibliography}{10}
\providecommand{\url}[1]{#1}
\csname url@rmstyle\endcsname
\providecommand{\newblock}{\relax}
\providecommand{\bibinfo}[2]{#2}
\providecommand\BIBentrySTDinterwordspacing{\spaceskip=0pt\relax}
\providecommand\BIBentryALTinterwordstretchfactor{4}
\providecommand\BIBentryALTinterwordspacing{\spaceskip=\fontdimen2\font plus
\BIBentryALTinterwordstretchfactor\fontdimen3\font minus
  \fontdimen4\font\relax}
\providecommand\BIBforeignlanguage[2]{{%
\expandafter\ifx\csname l@#1\endcsname\relax
\typeout{** WARNING: IEEEtran.bst: No hyphenation pattern has been}%
\typeout{** loaded for the language `#1'. Using the pattern for}%
\typeout{** the default language instead.}%
\else
\language=\csname l@#1\endcsname
\fi
#2}}

\bibitem{Rekleitis1998}
I.~M. Rekleitis, G.~Dudek, and E.~E. Milios, ``On multiagent exploration,'' in
  \emph{Vision Interface}, 1998, pp. 455--461.

\bibitem{quattrinili2016iser-vo}
A.~{Quattrini Li}, A.~Coskun, S.~M. Doherty, S.~Ghasemlou, A.~S. Jagtap,
  M.~Modasshir, S.~Rahman, A.~Singh, M.~Xanthidis, J.~M. O'Kane, and
  I.~Rekleitis, ``Experimental comparison of open source vision based state
  estimation algorithms,'' in \emph{Proc. ISER}, 2016, pp. 775--786.

\bibitem{JoshiIROS2019}
B.~Joshi, S.~Rahman, M.~Kalaitzakis, B.~Cain, J.~Johnson, M.~Xanthidis,
  N.~Karapetyan, A.~Hernandez, A.~{Quattrini Li}, N.~Vitzilaios, and
  I.~Rekleitis, ``{Experimental Comparison of Open Source Visual-Inertial-Based
  State Estimation Algorithms in the Underwater Domain},'' in \emph{Proc.
  IROS}, 2019, pp. 7221--7227.

\bibitem{ShkurtiIROS2017}
F.~Shkurti, W.~Chang, P.~Henderson, M.~Islam, J.~Higuera, J.~Li, T.~Manderson,
  A.~Xu, G.~Dudek, and J.~Sattar, ``Underwater multi-robot convoying using
  visual tracking by detection,'' in \emph{Proc. IROS}, 2017, pp. 4189--4196.

\bibitem{Manderson2017jfr}
T.~Manderson, J.~Li, N.~Dudek, D.~Meger, and G.~Dudek, ``{Robotic Coral Reef
  Health Assessment Using Automated Image Analysis},'' \emph{J. Field Robot.},
  vol.~34, no.~1, pp. 170--187, 2017.

\bibitem{Manderson2020rss}
T.~Manderson, J.~C. Gamboa, S.~Wapnick, J.-F. Tremblay, D.~Meger, and G.~Dudek,
  ``Vision-based goal-conditioned policies for underwater navigation in the
  presence of obstacles,'' in \emph{Proc. RSS}, July 2020.

\bibitem{lowe1999object}
D.~G. Lowe, ``Object recognition from local scale-invariant features,'' in
  \emph{Proc. ICCV}, vol.~2, 1999, pp. 1150--1157.

\bibitem{rothganger20063d}
F.~Rothganger, S.~Lazebnik, C.~Schmid, and J.~Ponce, ``{3D object modeling and
  recognition using local affine-invariant image descriptors and multi-view
  spatial constraints},'' \emph{Int. J. Comput. Vision}, 2006.

\bibitem{collet2011moped}
A.~Collet, M.~Martinez, and S.~S. Srinivasa, ``The {MOPED} framework: Object
  recognition and pose estimation for manipulation,'' \emph{Int. J. Robot.
  Res.}, vol.~30, no.~10, pp. 1284--1306, 2011.

\bibitem{pose_tracking_ismar}
D.~{Wagner}, G.~{Reitmayr}, A.~{Mulloni}, T.~{Drummond}, and D.~{Schmalstieg},
  ``Pose tracking from natural features on mobile phones,'' in \emph{IEEE/ACM
  Int. Symp. on Mix. and Aug. Reality}, 2008, pp. 125--134.

\bibitem{ssd_6d}
W.~Kehl, F.~Manhardt, F.~Tombari, S.~Ilic, and N.~Navab, ``{SSD-6D}: Making
  {RGB}-based 3{D} detection and 6{D} pose estimation great again,'' in
  \emph{Proc. ICCV}, 2017.

\bibitem{Sundermeyer_2018_ECCV}
M.~Sundermeyer, Z.-C. Marton, M.~Durner, M.~Brucker, and R.~Triebel, ``Implicit
  3d orientation learning for 6d object detection from rgb images,'' in
  \emph{ECCV}, 2018.

\bibitem{xiang2017posecnn}
Y.~Xiang, T.~Schmidt, V.~Narayanan, and D.~Fox, ``{PoseCNN}: A convolutional
  neural network for {6D} object pose estimation in cluttered scenes,''
  \emph{Proc. RSS}, 2018.

\bibitem{bb8}
M.~Rad and V.~Lepetit, ``{BB8:} a scalable, accurate, robust to partial
  occlusion method for predicting the 3d poses of challenging objects without
  using depth,'' in \emph{Proc. ICCV}, 2017.

\bibitem{single6dpose}
B.~Tekin, S.~N. Sinha, and P.~Fua, ``Real-time seamless single shot 6d object
  pose prediction,'' in \emph{Proc. CVPR}, 2018.

\bibitem{segpose}
Y.~Hu, J.~Hugonot, P.~Fua, and M.~Salzmann, ``Segmentation-driven 6d object
  pose estimation,'' in \emph{Proc. CVPR}, 2019.

\bibitem{linemod}
S.~Hinterstoisser, V.~Lepetit, S.~Ilic, S.~Holzer, G.~Bradski, K.~Konolige, and
  N.~Navab, ``Model based training, detection and pose estimation of
  texture-less 3{D} objects in heavily cluttered scenes,'' in \emph{ACCV},
  2013, pp. 548--562.

\bibitem{occluded_linemod}
A.~Krull, E.~Brachmann, F.~Michel, M.~Ying~Yang, S.~Gumhold, and C.~Rother,
  ``Learning analysis-by-synthesis for 6d pose estimation in rgb-d images,'' in
  \emph{Proc. ICCV}, 2015.

\bibitem{DudekIROS2005}
G.~Dudek, M.~Jenkin, C.~Prahacs, A.~Hogue, J.~Sattar, P.~Giguere, A.~German,
  H.~Liu, S.~Saunderson, A.~Ripsman, S.~Simhon, L.~A. Torres-Mendez, E.~Milios,
  P.~Zhang, and I.~Rekleitis, ``A visually guided swimming robot,'' in
  \emph{Proc. IROS}, 2005.

\bibitem{manderson2018aqua}
T.~Manderson, I.~Karp, and G.~Dudek, ``Aqua underwater simulator,'' in
  \emph{Proc. IROS}, 2018.

\bibitem{CycleGAN2017}
J.-Y. Zhu, T.~Park, P.~Isola, and A.~A. Efros, ``Unpaired image-to-image
  translation using cycle-consistent adversarial networks,'' in \emph{Proc.
  ICCV}, 2017.

\bibitem{yolov3}
J.~Redmon and A.~Farhadi, ``{YOLO}v3: An incremental improvement,'' 2018.

\bibitem{epnp}
V.~Lepetit, F.~Moreno-Noguer, and P.~Fua, ``{{EP}n{P:} An accurate {O}(n)
  solution to the {P}n{P} problem},'' \emph{Int. Journal of Comp. Vision},
  2009.

\bibitem{Brachmann2014}
E.~Brachmann, A.~Krull, F.~Michel, S.~Gumhold, J.~Shotton, and C.~Rother,
  \emph{Learning 6D Object Pose Estimation Using 3D Object Coordinates}.\hskip
  1em plus 0.5em minus 0.4em\relax Springer, 2014, vol. 8690.

\bibitem{CHOI2016595}
C.~Choi and H.~I. Christensen, ``{RGB-D object pose estimation in unstructured
  environments},'' \emph{Robotics and Aut. Systems}, vol.~75, pp. 595 -- 613,
  2016.

\bibitem{Sock2017Multiview6O}
J.~Sock, S.~H. Kasaei, L.~S. Lopes, and T.-K. Kim, ``Multi-view 6d object pose
  estimation and camera motion planning using rgbd images,'' \emph{IEEE Int.
  Conf. on Comp. Vision Workshops (ICCVW)}, 2017.

\bibitem{kostas_6D}
G.~{Pavlakos}, X.~{Zhou}, A.~{Chan}, K.~G. {Derpanis}, and K.~{Daniilidis},
  ``{6-DoF object pose from semantic keypoints},'' in \emph{Proc. ICRA}, 2017,
  pp. 2011--2018.

\bibitem{sift}
D.~Lowe, ``Distinctive image features from scale-invariant keypoints,''
  \emph{IJCV}, 11 2004.

\bibitem{NIPS2012_4848}
T.~Trzcinski, M.~Christoudias, V.~Lepetit, and P.~Fua, ``Learning image
  descriptors with the boosting-trick,'' in \emph{Proc. NeurIPS}, 2012.

\bibitem{Wohlhart_2015_CVPR}
P.~Wohlhart and V.~Lepetit, ``{Learning Descriptors for Object Recognition and
  3D Pose Estimation},'' in \emph{Proc. CVPR}, 2015.

\bibitem{doumanoglou2016siamese}
A.~Doumanoglou, V.~Balntas, R.~Kouskouridas, and T.-K. Kim, ``Siamese
  regression networks with efficient mid-level feature extraction for 3{D}
  object pose estimation,'' 2016.

\bibitem{peng2019pvnet}
S.~Peng, Y.~Liu, Q.~Huang, X.~Zhou, and H.~Bao, ``{PVNet: Pixel-wise Voting
  Network for 6DoF Pose Estimation},'' in \emph{Proc. CVPR}, 2019.

\bibitem{Zakharov_2019_ICCV}
S.~Zakharov, I.~Shugurov, and S.~Ilic, ``{DPOD: 6D Pose Object Detector and
  Refiner},'' in \emph{Proc. ICCV}, 2019.

\bibitem{Gupta_2019_ICCV_Workshops}
K.~Gupta, L.~Petersson, and R.~Hartley, ``{CullNet: Calibrated and Pose Aware
  Confidence Scores for Object Pose Estimation},'' in \emph{ICCV Workshops},
  2019.

\bibitem{li2017deepim}
Y.~Li, G.~Wang, X.~Ji, Y.~Xiang, and D.~Fox, ``{DeepIM}: {Deep Iterative
  Matching} for {6D} pose estimation,'' in \emph{Proc. ECCV}, 2018.

\bibitem{Li_2019_ICCV}
Z.~Li, G.~Wang, and X.~Ji, ``{CDPN: Coordinates-Based Disentangled Pose Network
  for Real-Time RGB-Based 6-DoF Object Pose Estimation},'' in \emph{Proc.
  ICCV}, 2019.

\bibitem{ipose}
O.~Hosseini Jafari, S.~K. Mustikovela, K.~Pertsch, E.~Brachmann, and
  C.~Rother, ``i{P}ose: Instance-aware 6{D} pose estimation of partly occluded
  objects,'' in \emph{Computer Vision -- ACCV}, 2018, pp. 477--492.

\bibitem{Oberweger2018MakingDH}
M.~Oberweger, M.~Rad, and V.~Lepetit, ``Making deep heatmaps robust to partial
  occlusions for {3D} object pose estimation,'' \emph{CoRR}, vol.
  abs/1804.03959, 2018.

\bibitem{Rozantsev2019BeyondSW}
A.~Rozantsev, M.~Salzmann, and P.~Fua, ``Beyond sharing weights for deep domain
  adaptation,'' \emph{{IEEE} Trans. Pattern Anal. Mach. Intell.}, vol.~41, pp.
  801--814, 2019.

\bibitem{featuremapping}
M.~Rad, M.~Oberweger, and V.~Lepetit, ``Feature mapping for learning fast and
  accurate 3{D} pose inference from synthetic images,'' in \emph{Proc. CVPR},
  2018.

\bibitem{koreitem_oceans2018}
K.~Koreitem, J.~Li, I.~Karp, T.~Manderson, F.~Shkurti, and G.~Dudek,
  ``Synthetically trained 3d visual tracker of underwater vehicles,'' in
  \emph{MTS/IEEE OCEANS}, Charleston, SC, USA, Oct. 2018.

\bibitem{gan}
I.~Goodfellow, J.~Pouget-Abadie, M.~Mirza, B.~Xu, D.~Warde-Farley, S.~Ozair,
  A.~Courville, and Y.~Bengio, ``Generative adversarial nets,'' in \emph{Proc.
  NeurIPS}, 2014, pp. 2672--2680.

\bibitem{coco}
T.-Y. Lin, M.~Maire, S.~Belongie, J.~Hays, P.~Perona, D.~Ramanan,
  P.~Doll{\'a}r, and C.~L. Zitnick, ``{Microsoft {COCO}: Common objects in
  context},'' in \emph{Proc. ECCV}, 2014, pp. 740--755.

\bibitem{yolov1}
J.~Redmon, S.~Divvala, R.~Girshick, and A.~Farhadi, ``You only look once:
  Unified, real-time object detection,'' in \emph{Proc. CVPR}, 2016.

\bibitem{nonmaxsupression}
R.~Rothe, M.~Guillaumin, and L.~V. Gool, ``Non-maximum suppression for object
  detection by passing messages between windows,'' in \emph{ACCV}, 2014.

\bibitem{Brachmann_2016_CVPR}
E.~Brachmann, F.~Michel, A.~Krull, M.~Ying~Yang, S.~Gumhold, and c.~Rother,
  ``Uncertainty-driven 6{D} pose estimation of objects and scenes from a single
  {RGB} image,'' in \emph{Proc. CVPR}, 2016.

\bibitem{ModasshirCRV2018}
M.~Modasshir, A.~{Quattrini Li}, and I.~Rekleitis, ``Deep neural networks: a
  comparison on different computing platforms,'' in \emph{Proc. CRV}, 2018, pp.
  383--389.

\bibitem{RahmanIROS2019a}
S.~Rahman, A.~{Quattrini Li}, and I.~Rekleitis, ``{SVIn2: An Underwater SLAM
  System using Sonar, Visual, Inertial, and Depth Sensor},'' in \emph{Proc.
  IROS}, 2019, pp. 1861--1868.

\end{thebibliography}

\end{document}